\documentclass[]{xsparklab}

\usepackage[utf8]{inputenc}
\usepackage{csquotes}
\usepackage{amsmath,amssymb,amsfonts}
\usepackage{mathtools}
\usepackage{amsthm}
\usepackage{xspace}
\usepackage{enumitem}
\usepackage{graphicx}
\usepackage{booktabs}
\usepackage{geometry}
\usepackage{titlesec}
\usepackage{hyperref}
\usepackage{tabularx}
\usepackage{array}
\usepackage{graphicx}
\RequirePackage{gradient-text}

\titleformat*{\paragraph}{\rmfamily\bfseries}

\title{
\begin{center}
Towards Trustworthy Embodied Intelligence: A Systems Framework and Graded Trustworthiness Levels\\
{\large\mdseries\itshape From Robust and Safe Task Execution to Evidence-Based Deployment Governance}
\end{center}
}

\author{SparkLab@Xspark AI, THU, MMLab@HKU, PKU, HKUST (GZ)\protect\footnotemark}

\titleimage[
  width=\linewidth,
  height=8cm,
  keepaspectratio
]{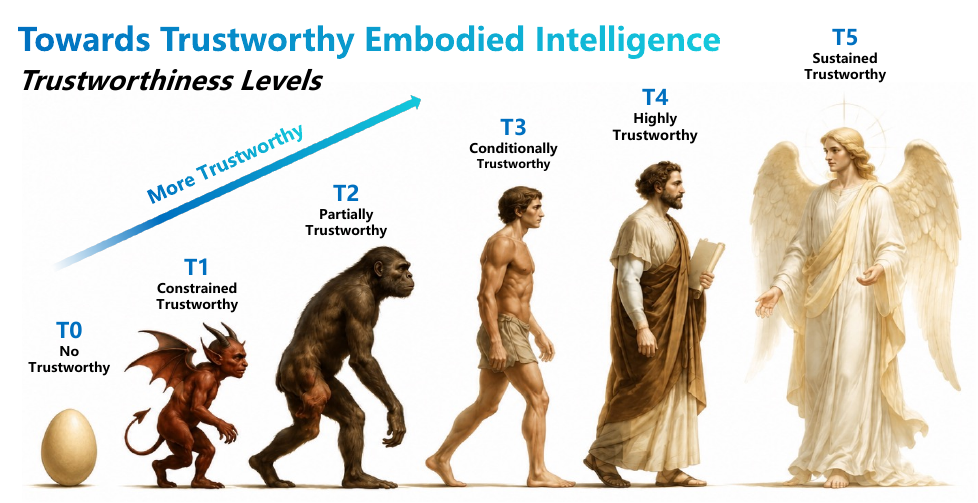}

\abstract{
Embodied intelligence integrates learned perception and decision making with real-time computation, control, and physical interaction. Because failures can cause immediate physical or operational harm, task completion alone does not establish trustworthiness. We define trustworthy embodied intelligence as the sustained capacity to execute specified tasks reliably under environmental and system variation while maintaining risk within acceptable bounds. We term this objective \emph{sustained safe success}. Its supporting mechanisms are organized into four interdependent layers. The \emph{model layer} generates task-competent action proposals with calibrated uncertainty and explicit safety preferences. The \emph{system layer} realizes authorized actions dependably through integrated sensing, computation, control, hardware safeguards, fault containment, and fallback. The \emph{evidence layer} substantiates bounded claims through evaluation, verification, validation, traceability, and structured assurance arguments. The \emph{deployment layer} maintains claim validity through runtime monitoring, authority management, intervention, incident response, and controlled updates. Because assumptions and failures propagate across these layers, neither model capability, isolated safeguards, nor benchmark performance alone can establish end-to-end trustworthiness. Drawing on embodied AI, robotics, control, dependable computing, distributed systems, and autonomous driving, we further propose a non-normative hierarchy of trustworthiness levels. This hierarchy grades the strength of bounded deployment claims across task capability, safety, system assurance, operational governance, and supporting evidence, providing a basis for bounded deployment, comparative evaluation, research prioritization, and future standardization.}

\metadata[Project Page]{\url{https://xsparkai.com/sparklab/towards-trustworthy-eai}
\\ \textbf{Published Date:} July 28, 2026}
\begin{document}

\maketitle
\footnotetext{See \hyperref[contributions]{Contributors} section for full author list. Please send correspondence to \href{mailto:sparklab@xsparkai.com}{sparklab@xsparkai.com}.}
\xsparktoc

\newpage

\section{Introduction}
\label{sec:introduction}

Recent advances in vision--language--action models, robot foundation models, and embodied multimodal policies~\citep{brohan2022rt,driess2023palm,brohan2023rt2,kim2024openvla,black2024pi0}, together with progress in world models~\citep{hafner2023dreamerv3}, tactile perception~\citep{tactile2025outlook}, and whole-body control~\citep{cheng2024wholebody}, have expanded the capabilities of embodied systems in manipulation, navigation, and human interaction. These developments extend artificial intelligence from digital inference to closed-loop physical interaction, thereby changing the consequences of error. In a digital system, an erroneous prediction may degrade information or service quality; in an embodied system, it may induce unsafe motion, unintended contact, equipment damage, or physical harm. The central question therefore shifts from whether an agent can complete a task to whether an integrated system can do so reliably under environmental and system variation while maintaining risk within acceptable bounds~\citep{kojima2025physicalrisk,li2026safetyembodied,ma2026breaks}.

Experimental capability alone does not justify deployment. Laboratory and simulation evaluations commonly constrain object sets, workspace conditions, task duration, communication, and human interaction. In deployment, however, systems encounter distribution shift~\citep{peng2017sim,tobin2017domainrand}, sensing and actuation degradation, communication delay, software change, and unanticipated human behavior~\citep{kojima2025physicalrisk,li2026safetyembodied}. This gap is especially consequential in long-horizon and contact-rich tasks: an early error in perception, grounding, or planning may propagate across subtasks and become apparent only after physical contact has occurred or recovery options have narrowed~\citep{kojima2025physicalrisk,kim2026longhorizon}. Performance under restricted evaluation conditions therefore does not establish suitability for the broader deployment domain.

We use \emph{trustworthy embodied intelligence} to denote \emph{sustained safe success}: reliable completion of intended tasks while physical, semantic, procedural, and operational risks remain within acceptable bounds. This objective requires two jointly necessary properties:
\begin{itemize}[leftmargin=*]
    \item \textbf{Task capability:} adequate task success, quality, efficiency, timeliness, robustness, and generalization over the intended task distribution;
    \item \textbf{Safety:} behavior consistent with appropriate safety preferences and explicit constraints on actions, consequences, and authority.
\end{itemize}

Safety preferences rank otherwise feasible behaviors by relative risk, whereas safety constraints define non-negotiable conditions for admissible execution. Capability and safety are not substitutable: high average task success may conceal rare but severe failures, whereas indiscriminate refusal precludes useful execution. Because acceptable performance and residual risk are application-specific, every trustworthiness claim is bounded to a particular system configuration, task distribution, embodiment, environment, authority structure, and body of evidence.

Sustained safe success also depends on properties beyond the learned model. \emph{Reliability} concerns the continuity and correctness of required service over time~\citep{avizienis2004dependability}; \emph{physical controllability} concerns whether available sensing, actuation, and control authority can preserve or recover admissible states~\citep{ames2019cbf}; and \emph{governability} concerns whether authorized humans or supervisory mechanisms can observe, constrain, interrupt, degrade, recover, or transfer execution. \emph{Evaluability} requires claims to be measurable, traceable, and conditioned on explicit assumptions, whereas \emph{deployability} requires those claims to remain valid through operation, maintenance, updates, and environmental change.

Trustworthiness is therefore an end-to-end property of a deployed system, not an attribute of an individual model, component, or benchmark score. Robust perception does not ensure safe planning; a stable controller may faithfully execute an unsafe objective; redundant components may share software, communication, calibration, or power dependencies; and formal guarantees hold only under their stated assumptions. Likewise, task-completion rate conflates safe success with success achieved through unsafe behavior. Evaluation must distinguish these outcomes rather than reward completion alone.

We organize these requirements into four interdependent layers. The \emph{model layer} generates task-relevant action proposals while representing uncertainty, safety preferences, and relevant constraints. The \emph{system layer} realizes admissible actions through dependable sensing, computation, control, actuation, hardware safeguards, fault containment, and fallback. The \emph{evidence layer} determines whether bounded trustworthiness claims are justified by evaluation, verification, validation, traceability, and structured assurance arguments. The \emph{deployment layer} maintains claim validity through runtime monitoring, authority management, intervention, incident response, and controlled updates. Section~\ref{sec:framework} develops the interactions among these layers.

\subsection{Scope and Perspective}
\label{subsec:scope_perspective}

This survey considers systems that close a perception--decision--action loop through physical hardware, including manipulators, mobile robots, humanoids, industrial and service robots, networked and multi-robot systems, and autonomous vehicles insofar as their engineering principles transfer to general embodied intelligence. It covers learned and conventional software; sensing and actuation; real-time computation and communication; control and system integration; physical human--robot interaction; fault-tolerant execution; evaluation; and deployment governance.

Purely digital model safety is considered only where it affects physical task interpretation, planning, capability selection, control, or authority. We include capability-oriented research where it establishes competence necessary for safe execution, reveals capability--safety interactions, or defines interfaces through which model decisions affect physical behavior. We assume neither universal performance thresholds nor a common acceptable level of residual risk across embodiments and application domains.

The relevant literature spans embodied AI, AI safety, robotics, control, cybersecurity, dependable computing, distributed systems, human factors, and engineering standards. Existing studies variously emphasize model robustness, runtime safeguards, fault-tolerant control, evaluation protocols, intervention, or lifecycle assurance~\citep{kojima2025physicalrisk,li2026safetyembodied,ma2026breaks,kim2026longhorizon,qin2026runtimegovernance,quamar2024faultdiagnosis}. We synthesize these perspectives according to their roles in establishing and maintaining bounded trustworthiness claims, addressing five questions:
\begin{enumerate}[leftmargin=*]
    \item How should task capability and safety be jointly defined and evaluated for embodied systems?
    \item How do faults, uncertainty, malicious inputs, timing violations, and invalid assumptions propagate across the model, system, evidence, and deployment layers?
    \item Which mechanisms preserve useful capability while preventing, constraining, detecting, and recovering from unsafe behavior?
    \item Which forms of formal, statistical, empirical, and operational evidence substantiate bounded deployment claims?
    \item How should mechanisms and evidence evolve from laboratory evaluation to constrained deployment and continuous field assurance?
\end{enumerate}

\subsection{Contributions}
\label{subsec:introduction_contributions}

This survey makes four contributions:
\begin{enumerate}[leftmargin=*]
    \item \textbf{A deployment-oriented definition of trustworthiness.}
    We define trustworthy embodied intelligence as sustained safe success and distinguish the primary objectives of capability and safety from the supporting properties of reliability, physical controllability, governability, evaluability, and deployability.

    \item \textbf{A four-layer systems framework.}
    We organize trustworthiness mechanisms into model, system, evidence, and deployment layers, specifying their respective responsibilities and the cross-layer interactions required to support end-to-end claims.

    \item \textbf{A cross-layer synthesis of mechanisms, failures, and evidence.}
    We connect embodied AI with robotics, control, dependable computing, distributed systems, cybersecurity, and autonomous driving, characterizing failure propagation across layer boundaries and organizing existing mechanisms by their roles in prevention, containment, detection, recovery, validation, and governance.

    \item \textbf{A graded hierarchy of Trustworthy Embodied Intelligence.}
    We propose a non-normative T0--T5 hierarchy, ranging from no substantiated trustworthiness to sustained trustworthiness under governed change. Each level characterizes a progressively stronger bounded claim across capability, safety, system assurance, evidence, and governance, and applies to a specified system, task, embodiment, environment, authority structure, and evidence base. The hierarchy supports comparative evaluation, bounded deployment, research prioritization, and future standardization rather than certification.
\end{enumerate}

\subsection{Organization}
\label{subsec:paper_organization}

Section~\ref{sec:why_trustworthy} analyzes the consequences of insufficient trustworthiness at each layer. Section~\ref{sec:framework} presents the four-layer framework and its cross-layer coordination. Sections~\ref{sec:model_layer}, \ref{sec:system_layer}, \ref{sec:evidence_layer}, and~\ref{sec:deployment_layer} review the responsibilities, representative mechanisms, and limitations of the model, system, evidence, and deployment layers. Section~\ref{sec:levels_deployment} introduces the T0--T5 levels of Trustworthy Embodied Intelligence; Section~\ref{sec:prior_work_summary} synthesizes prior work across the four layers; and Section~\ref{sec:conclusion} concludes.
\section{Why Trustworthiness Is Necessary}
\label{sec:why_trustworthy}

\begin{figure*}[t]
    \centering
    \includegraphics[width=\textwidth]{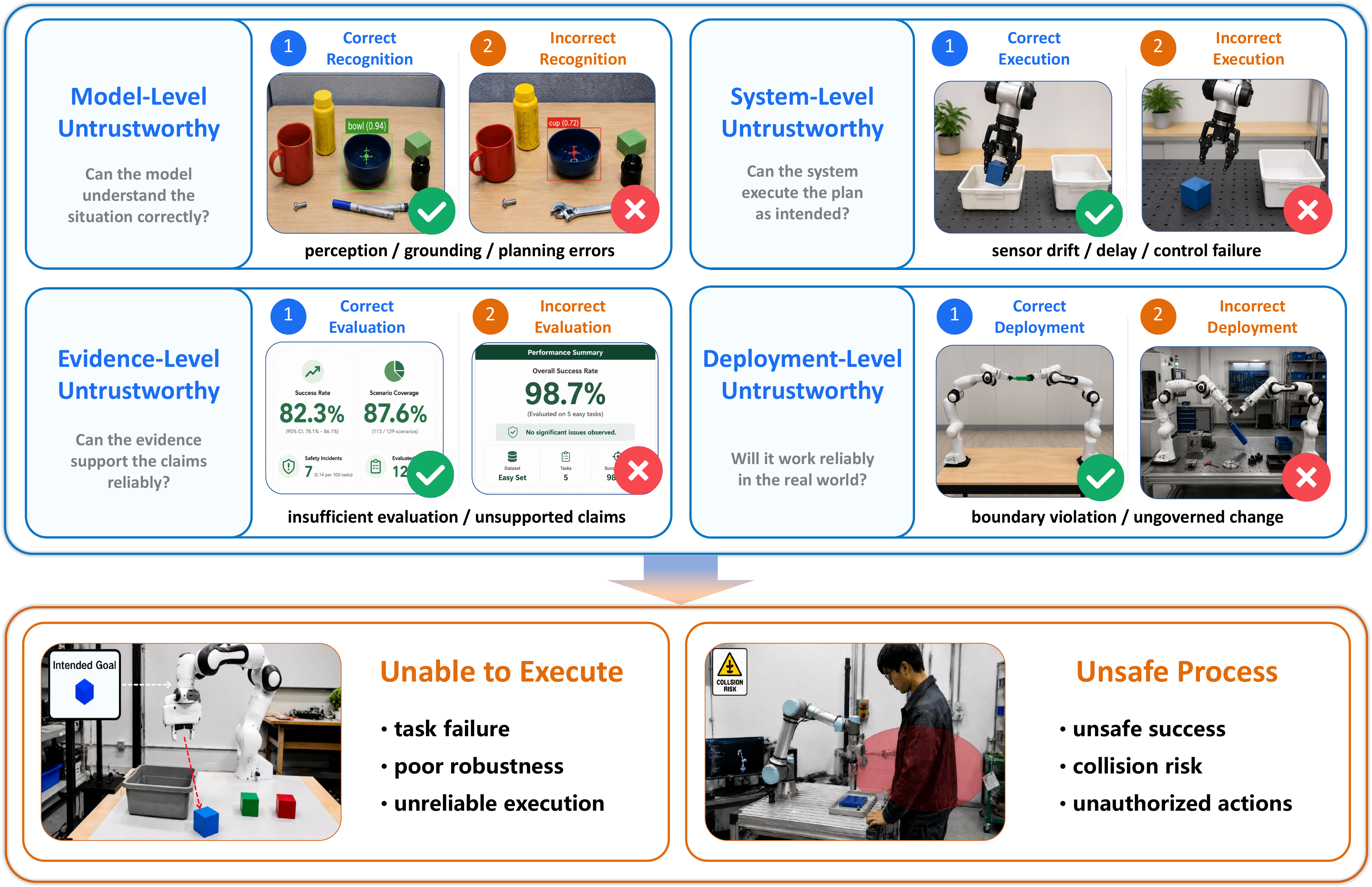}
    \caption{Potential consequences of untrustworthy embodied intelligence, including physical safety risks, privacy violations, security threats, social impacts, and accountability failures.}
    \label{fig:untrustworthy_harms}
\end{figure*}

Embodied systems translate computational outputs into physical action. Errors in perception, reasoning, communication, control, or actuation can therefore cause not only task failure but also unsafe contact, equipment damage, operational disruption, and physical harm, some of which is irreversible~\citep{kojima2025physicalrisk,li2026safetyembodied,ma2026breaks}. Nor does a universally safe response exist: immediate stopping can itself be hazardous when a robot is balancing, bearing a load, or engaged in physical interaction, and the appropriate response may instead be a task-dependent minimum-risk transition~\citep{haddadin2017collisions,zanella2026phri}. Sustaining useful task execution at acceptable residual risk throughout deployment therefore requires trustworthiness as defined in Section~\ref{sec:introduction}, and this section examines the consequences of its absence at each layer.

\subsection{Consequences of Layer-Specific Deficiencies}
\label{subsec:layer_specific_deficiencies}

\paragraph{Model layer.}
Deficiencies in perception, grounding, planning, uncertainty estimation, or safety alignment can generate ineffective, infeasible, unauthorized, or unsafe action proposals. Calibration and out-of-distribution detection can expose some epistemic limitations, but their effectiveness depends on whether uncertainty is represented and acted upon downstream~\citep{guo2017calibration,hendrycks2017ood}. In long-horizon tasks, an early error may propagate across subtasks and become apparent only when recovery is difficult or infeasible~\citep{kim2026longhorizon}. Capability alone is therefore insufficient: an otherwise competent policy may act on incomplete state estimates, invalid assumptions, or under-specified safety constraints~\citep{ahn2022saycan,amodei2016concrete}.

\paragraph{System layer.}
System deficiencies can transform an otherwise admissible proposal into unsafe physical behavior. Sensor drift, stale state estimates, communication delay, actuator degradation, control saturation, and insufficient braking authority can invalidate the assumptions under which an action was proposed. Without timely fault detection, local protection, and fallback, these failures may propagate to unacceptable physical consequences~\citep{avizienis2004dependability,ames2019cbf,gao2015faultsurvey}.

\paragraph{Evidence layer.}
Evidence deficiencies permit deployment claims that exceed what evaluation establishes. Aggregate performance measures can obscure rare failures, unsafe intermediate behavior, and scenario-dependent limitations~\citep{corso2021survey}. Simulation is bounded by model fidelity and the validity of its assumptions, whereas physical testing is bounded by scenario coverage, exposure, and test duration~\citep{tobin2017domainrand,riedmaier2020scenario,kalra2016driving}. Evidence established for one embodiment, task distribution, or configuration therefore cannot be presumed valid after a material change.

\paragraph{Deployment layer.}
Deployment deficiencies allow a previously supported claim to lapse during operation. Changes in hardware, software, calibration, tools, communication, human behavior, or task scope can alter both capability and risk. Monitoring and human intervention provide limited protection when deviations are detected too late or when operators lack the information, time, or authority required to intervene effectively~\citep{qin2026runtimegovernance,santonidesio2018mhc}.

\subsection{Cross-Layer Failure Propagation}
\label{subsec:cross_layer_failure_propagation}

The four layers are interdependent. System safeguards may contain a model error~\citep{sha2001simplex}, but harm may still occur when sensing is degraded or protection is unavailable. Conversely, dependable control cannot determine whether a high-level objective is semantically appropriate or authorized. Evidence supports deployment only under stated assumptions, which deployment mechanisms must preserve throughout use.

This interdependence gives rise to three recurrent problems. The \emph{semantic--physical gap} arises when abstract instructions omit relevant geometric, dynamic, or contact constraints. The \emph{action--consequence gap} arises because the same commanded action can produce different physical consequences depending on the current state, so admissibility cannot be judged from the command alone. \emph{Cross-layer non-compositionality} arises when individually capable, reliable, or verified components fail to form a trustworthy system because their assumptions, interfaces, or constraints are incompatible~\citep{leveson2011safer}.

No individual layer can establish end-to-end trustworthiness. The next section formalizes the complementary responsibilities of the four layers and the interactions through which they support bounded trustworthiness claims.
\section{A Four-Layer Framework for Trustworthy Embodied Intelligence}
\label{sec:framework}

Trustworthy embodied intelligence requires more than capable models or isolated safety mechanisms: it requires an architecture that connects task capability, dependable physical execution, evidence for bounded claims, and governance throughout deployment. We develop this architecture by treating autonomous driving as an engineering analogue and abstracting the assurance principles that transfer to embodied intelligence.

\subsection{Lessons from Trustworthy Autonomous Driving}
\label{subsec:driving_analogy}

Autonomous driving provides a useful analogue because both domains combine learned perception and decision making with real-time control, physical interaction, and operation under uncertainty. In autonomous driving, trustworthiness is not inferred from driving performance alone. It depends on specified operating conditions and safety requirements, scenario-based evaluation, validation in representative conditions, fallback behavior, and continued operational assurance~\citep{riedmaier2020scenario,iso21448,goodloe2022assuring}.

\begin{figure*}[t]
    \centering
    \includegraphics[width=\textwidth]{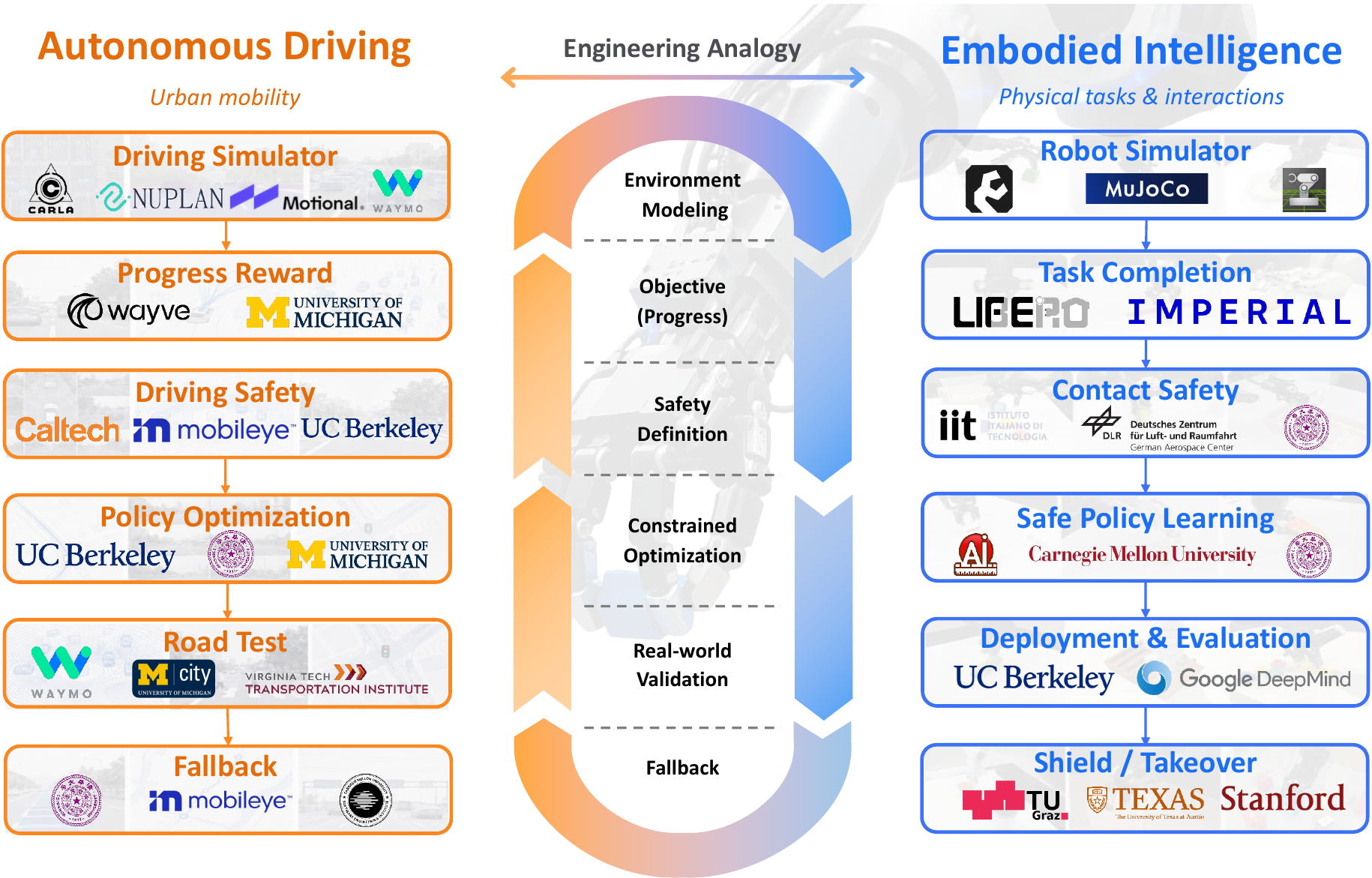}
    \caption{Engineering analogy between autonomous driving and trustworthy embodied intelligence across modeling, objectives, safety, optimization, validation, and fallback.}
    \label{fig:autonomous_driving_analogy}
\end{figure*}

Figure~\ref{fig:autonomous_driving_analogy} illustrates the resulting engineering correspondence. Driving and robot simulators provide controlled environments for modeling and evaluation; progress-oriented driving and task-completion objectives define useful behavior; vehicle-safety and contact-safety constraints bound physical risk; constrained optimization and safe learning shape behavior within these constraints; road testing and robot deployment provide real-world evidence; and fallback mechanisms mitigate the loss of nominal capability.

The analogy is informative but incomplete. General-purpose embodied systems must interpret open-ended instructions, interact with objects of diverse and uncertain properties, and manage changing contact conditions; as discussed in Section~\ref{sec:why_trustworthy}, they also lack a universally safe stopping response. Autonomous driving therefore provides an engineering reference, rather than a direct template, for trustworthy embodied intelligence.

\subsection{Framework Design}
\label{subsec:framework_design}

\begin{figure*}[t]
    \centering
    \includegraphics[width=\textwidth]{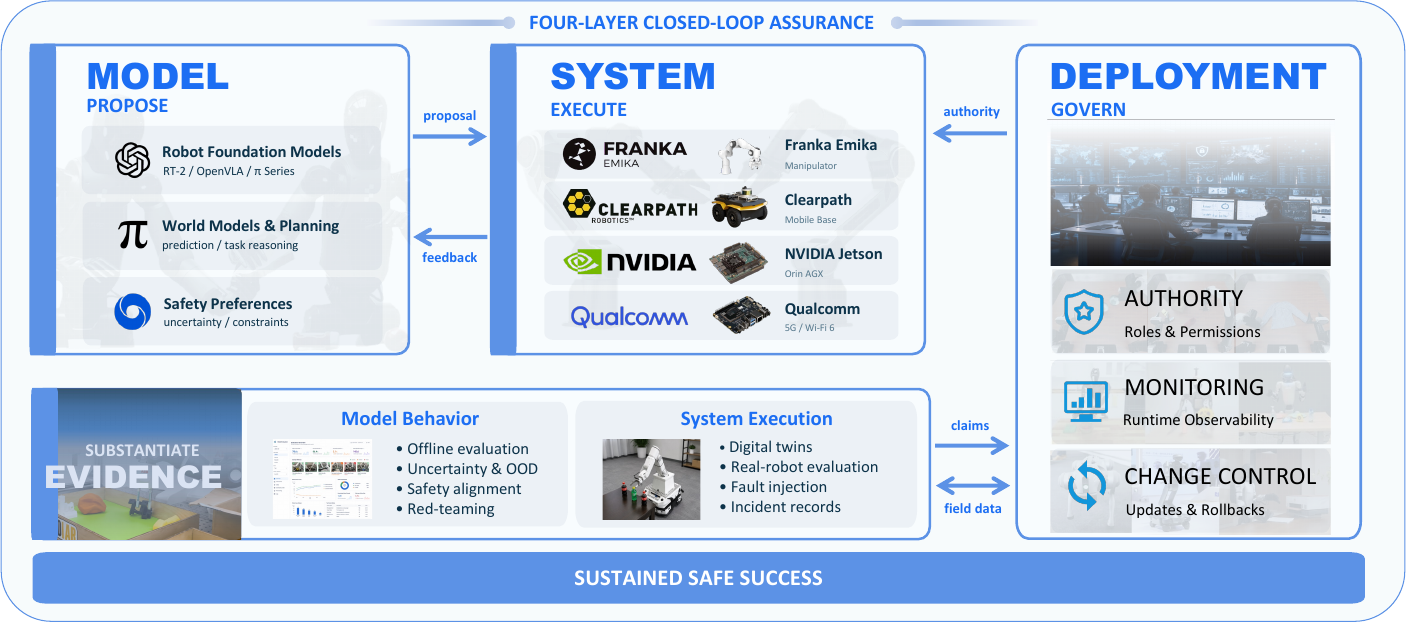}
    \caption{The four-layer framework for trustworthy embodied intelligence. The model layer proposes actions; the system layer realizes and constrains them physically; the evidence layer justifies bounded trustworthiness claims; and the deployment layer maintains claim validity during operation. Arrows denote the cross-layer exchange of proposals, constraints, uncertainty, system health, and evidence.}
    \label{fig:four_layer_architecture}
\end{figure*}

The analogy reveals a common principle: trustworthy physical autonomy requires coordinated support for decision making, physical execution, evidence, and runtime governance. We organize these functions into the four interdependent layers shown in Figure~\ref{fig:four_layer_architecture}.

Each layer holds a distinct responsibility. The \emph{model layer} generates task-competent action proposals with calibrated uncertainty and explicit safety preferences (Section~\ref{sec:model_layer}). The \emph{system layer} realizes authorized proposals through dependable sensing, computation, control, and actuation, and contains residual failures through local protection and fallback (Section~\ref{sec:system_layer}). The \emph{evidence layer} determines which bounded claims about the integrated system are justified by formal, statistical, empirical, and operational evidence (Section~\ref{sec:evidence_layer}). The \emph{deployment layer} maintains the validity of those claims during use through admission, monitoring, intervention, incident response, and change control (Section~\ref{sec:deployment_layer}).

The layers are functional responsibilities rather than software modules, and no layer is self-sufficient. Proposals flow downward only together with their preconditions, uncertainty, and active constraints; execution state, system health, and timing validity flow upward so that decisions are not made on stale assumptions. The evidence layer evaluates the model and system layers as an integrated whole, and the deployment layer enforces the assumptions under which that evidence was established. The object coordinating all four layers is the bounded trustworthiness claim introduced in Section~\ref{sec:introduction}: the model and system layers realize the claimed behavior, the evidence layer substantiates the claim, and the deployment layer preserves its validity. Sections~\ref{sec:model_layer} through~\ref{sec:deployment_layer} examine each layer in turn.
\section{Model-Level Trustworthiness}
\label{sec:model_layer}

The model layer transforms observations, task requests, and context into state representations, plans, skills, and action proposals. It includes learned and conventional components for perception, grounding, reasoning, planning, world modeling, and policy generation. Recent vision--language--action and physically grounded vision--language models expand this capability space~\citep{kim2024openvla,gao2024physically}, while work on verifiable instruction following illustrates the need to connect high-level task interpretation with executable constraints~\citep{quartey2025verifiably}.

Model-level trustworthiness concerns whether proposed behavior is effective, physically feasible, uncertainty-aware, safety-constrained, and consistent with applicable authority. This layer alone cannot establish end-to-end trustworthiness: a calibrated model may rely on degraded sensing, incomplete constraints, or inaccurate dynamics, and a safety-aligned policy may encounter conditions outside its supported distribution. Its role is therefore to reduce unsafe or ineffective proposals and to expose sufficient information for downstream system and deployment mechanisms to assess their admissibility.

\subsection{Grounded Perception and State Representation}
\label{subsec:grounded_perception}

Embodied perception must estimate task-relevant physical state rather than merely recognize semantic categories. Relevant quantities include object pose and motion, free space, support relations, contact state, material properties, human behavior, and the robot's own configuration. Visuo-tactile methods further illustrate the importance of contact-relevant state for fine-grained manipulation~\citep{huang20243d}. Errors in these quantities can impair task performance or induce physical risk even when object recognition is nominally correct.

Perception must operate under illumination variation, occlusion, transparency, motion blur, contamination, calibration drift, and distribution shift. It must also resist adversarial inputs and training-time compromise, including physical patches, spatiotemporal attacks on embodied agents, and backdoor triggers~\citep{brown2017patch,eykholt2018robust,liu2020spatiotemporal,gu2017badnets}. Robustness should therefore be assessed by its effects on downstream task execution and safety, rather than by classification accuracy alone~\citep{li2026safetyembodied,ma2026breaks}.

Multimodal sensing can improve observability but does not automatically provide independent evidence. Camera, depth, tactile, and proprioceptive channels may share calibration, timing, environmental, computational, or training-data dependencies. Fusion should consequently preserve source provenance, freshness, uncertainty, and disagreement rather than collapse inputs into an unqualified estimate.

Uncertainty is useful only when it changes decisions. Confidence estimates, evidential uncertainty, freshness indicators, and out-of-distribution signals~\citep{guo2017calibration,hendrycks2017ood,amini2020deep} should propagate to planning and execution. Excessive uncertainty may then trigger re-observation, viewpoint change, tactile verification, reduced motion, clarification, or deferral.

\subsection{Grounding and Safety-Aware Reasoning}
\label{subsec:grounding_reasoning}

Grounding connects task semantics to the current environment, embodiment, and authority structure. A linguistically valid instruction may nevertheless be physically infeasible, under-specified, or unauthorized. The model must therefore identify the intended entities and goals, assess available capabilities, and preserve relevant semantic, procedural, and physical constraints during task decomposition.

Affordance and capability reasoning can reject actions unsupported by the current embodiment or environment~\citep{ahn2022saycan,mees2022grounding}, and instructions can be grounded into executable programs whose primitives bound the available behavior~\citep{liang2023codeaspolicies}. However, learned affordances may reflect correlations rather than reliable physical constraints. Grounding should preserve uncertainty and distinguish observed, inferred, and assumed state. When an instruction omits safety-relevant information, the model should seek clarification or propose a restricted alternative rather than silently instantiate uncertain constraints.

Instruction integrity belongs to the same trust boundary. Prompt injection, conflicting commands, deceptive environmental content, malicious tool outputs, and physically triggered backdoors can alter task interpretation or action selection~\citep{greshake2023injection,zou2023universal,zhang2024badrobot,zhou2025goal}. Trusted policy and authority constraints should remain separate from untrusted task content. Moreover, a single policy must govern every executable channel: a textual refusal is ineffective if an action, code, or tool-use channel can still initiate prohibited behavior.

\subsection{Planning and Policy Generation}
\label{subsec:model_planning_policy}

Long-horizon tasks involve ordering constraints, intermediate state changes, resource dependencies, human interaction, and recovery decisions. An early grounding or planning error may remain latent until a later contact-rich step, while apparent task completion can conceal unsafe intermediate behavior~\citep{kim2026longhorizon}. Planning must therefore address procedural validity and recoverability, rather than endpoint success alone.

Task decomposition should ensure that each subtask's postconditions establish the preconditions of subsequent subtasks. Task-and-motion planning links symbolic decisions to kinematic reachability, collision avoidance, grasp stability, and resource feasibility~\citep{garrett2021tamp,ding2023task}. Temporal and spatial specifications can further encode ordering constraints, invariants, and prohibited states, although their guarantees remain conditional on specification completeness and state-estimate validity.

Policy learning shapes the distribution of proposed actions. Constrained reinforcement learning separates task utility from safety cost and optimizes performance subject to an explicit risk budget~\citep{achiam2017cpo,tessler2018reward,garcia2015saferl}. Related mechanisms include safety-aware preference alignment, action-space restrictions, certificate-guided policies, constrained generative planning, and predictive safety filters~\citep{ames2019cbf,tian2025safety,zhang2026constrained,yu2023safe,wabersich2021psf}. When such mechanisms only filter model outputs, they remain model-level mechanisms; independent enforcement during physical execution belongs to the system layer.

Constrained learning is limited by its objectives and safety signals. Scalar penalties may underrepresent rare severe events, be dominated by task reward, or induce excessive refusal. Evaluation should therefore distinguish safe success, unsafe success, intervention burden, efficiency, and availability rather than report aggregate reward alone.

\subsection{World Models and Long-Horizon Risk}
\label{subsec:model_world_models}

World models learn to predict task-relevant state evolution under candidate actions, supporting counterfactual evaluation, policy learning, and planning without executing every candidate on physical hardware~\citep{hafner2023dreamerv3}. For trustworthy embodied use, predicted state should represent not only task progress but also uncertainty, contact, constraint violations, and relevant failure conditions. Learned closed-loop sensor simulators can likewise support predictive evaluation, but require the same scrutiny of validity under deployment conditions~\citep{yang2023unisim}.

A visually plausible prediction need not be physically valid. Latent dynamics may omit friction, force, deformability, actuator limits, communication delay, or hardware degradation; prediction errors can then accumulate with planning horizon and induce trajectories that appear safe in simulation but fail physically. World-model predictions should therefore be calibrated against physical trajectories and repeatedly corrected using new observations.

SafeDojo illustrates the separation of capability and safety objectives in world-model-based policy learning: imagined trajectories receive distinct task-progress and collision signals, and optimization is adjusted when predicted safety cost exceeds a specified budget~\citep{tang2026safedojo}. Its supported claims remain bounded by the accuracy, coverage, and calibration of the world model and risk signal.

Long-horizon reasoning must also account for recoverability. An action with low immediate risk may lead to a state from which safe continuation is infeasible. Recovery-aware planning should preserve stable intermediate states, reversible actions, feasible stopping conditions, and alternative paths. It should also bound repeated retries, since individually minor failures can accumulate contact, wear, delay, and human burden.

\subsection{Model Integrity and Governable Interfaces}
\label{subsec:model_integrity_interfaces}

Training data shape both capability and safety. Datasets dominated by successful demonstrations may omit hazards, justified refusal, near misses, intervention, and recovery. Poisoning, mislabeling, and backdoors can further implant behavior triggered by specific objects, patterns, instructions, or conditions~\citep{gu2017badnets,zhou2025goal,biggio2012poisoning}. Data provenance should therefore record embodiment, sensing configuration, task conditions, operator involvement, interventions, failures, and known coverage limitations.

Pretrained backbones, adapters, reward and cost models, external tools, and generated code introduce dependencies whose changes can alter action distributions, confidence, latency, or safety behavior. Model artifacts should consequently be identifiable, versioned, and traceable. The evidence and deployment layers must then determine whether a change requires targeted regression testing or broader revalidation.

Model outputs should use structured, governable interfaces rather than unrestricted text, code, or low-level commands. A proposal should declare its target, reference frame, preconditions, expected effects, uncertainty, active constraints, completion criteria, timeout, and recovery options. Such interfaces allow the system layer to approve, restrict, modify, or reject proposals without interpreting opaque internal representations, and allow deployment governance to enforce policy-constrained execution~\citep{qin2026runtimegovernance,quartey2025verifiably}.

Calibrated deferral is itself a capability. When observations, permissions, or confidence do not justify a proposal, the model should gather information, request clarification, select a restricted alternative, or seek assistance~\citep{ren2023robots}. Insufficient deferral converts uncertainty into physical risk, whereas excessive deferral reduces task success and availability.

\begin{table*}[t]
\centering
\small
\setlength{\tabcolsep}{4pt}
\renewcommand{\arraystretch}{1.25}
\caption{Contributions and limitations of representative model-layer functions. A checkmark denotes a primary contribution, a triangle denotes a supporting contribution, and a dash denotes no direct contribution.}
\label{tab:model_layer_mechanisms}

\begin{tabularx}{\textwidth}{
    >{\raggedright\arraybackslash}p{0.18\textwidth}
    *{5}{>{\centering\arraybackslash}p{0.095\textwidth}}
    >{\raggedright\arraybackslash}X}
\toprule
\textbf{Model function}
&
\textbf{State and uncertainty}
&
\textbf{Grounding and constraints}
&
\textbf{Feasibility and planning}
&
\textbf{Risk-aware actions}
&
\textbf{Traceability and governance}
&
\textbf{Principal limitation}
\\
\midrule

Perception and state estimation
& $\checkmark$
& $\triangle$
& $\triangle$
& $\triangle$
& --
& Uncertainty may be miscalibrated or lost across interfaces.
\\

Grounding and reasoning
& $\triangle$
& $\checkmark$
& $\triangle$
& $\triangle$
& $\triangle$
& Physical or authority constraints may remain incomplete.
\\

Planning
& $\triangle$
& $\checkmark$
& $\checkmark$
& $\triangle$
& $\triangle$
& Guarantees depend on state, dynamics, and specification assumptions.
\\

Policy generation
& $\triangle$
& $\triangle$
& $\triangle$
& $\checkmark$
& $\triangle$
& Learned objectives may omit rare hazards or cause excessive refusal.
\\

World modeling
& $\checkmark$
& $\triangle$
& $\checkmark$
& $\checkmark$
& --
& Prediction errors and omitted dynamics accumulate with horizon.
\\

Model integrity and interfaces
& $\triangle$
& $\triangle$
& --
& $\triangle$
& $\checkmark$
& Integrity controls do not ensure safe integrated execution.
\\

\bottomrule
\end{tabularx}
\end{table*}

\subsection{Open Research Directions}
\label{subsec:model_open_directions}

\paragraph{Grounded safety specifications.}
Translating abstract instructions and safety preferences into embodiment-specific geometric, dynamic, procedural, and authority constraints while preserving semantic intent.

\paragraph{Actionable uncertainty.}
Developing uncertainty representations that remain calibrated across perception, planning, and policy interfaces and trigger proportionate information gathering, restriction, or deferral.

\paragraph{Long-horizon safe success.}
Defining objectives that capture cumulative risk, intermediate violations, recoverability, repeated attempts, and delayed consequences rather than local action safety or final completion alone.

\paragraph{Assurance under model adaptation.}
Detecting behaviorally relevant changes induced by fine-tuning, online adaptation, tool integration, or model replacement; localizing their assurance impact; and enabling efficient regression testing without assuming that safety transfers across versions.
\section{System-Level Trustworthiness}
\label{sec:system_layer}

The system layer realizes model-generated proposals through sensing, computation, communication, control, actuation, and physical interaction. Hardware is its physical assurance substrate rather than a separate layer. This layer determines whether the integrated perception--decision--action loop can execute authorized tasks reliably, preserve safety constraints, and contain residual model or component failures before they produce unacceptable consequences.

The system layer provides immediate execution safety through physical integration, local protection, fault containment, and fallback. In contrast, the deployment layer governs operational authority, human intervention, incident management, and lifecycle change. This distinction separates time-critical physical protection from broader runtime governance.

\subsection{Hardware as the Physical Assurance Substrate}
\label{subsec:hardware_substrate}

Hardware determines which physical states are observable, which actions are feasible, which timing requirements can be met, and which degraded conditions remain controllable. Its trustworthiness must therefore be assessed for the configured system, task, and environment. Component reliability alone is insufficient because sensor placement, shared dependencies, payloads, tools, thermal conditions, and system integration jointly affect both capability and risk.

\paragraph{Sensing and physical observability.}
Sensors determine whether task state, hazards, contact, and degradation are detected with sufficient accuracy and lead time. Relevant failures include bias, drift, saturation, dropout, contamination, misalignment, timestamp error, and calibration change. Gradual faults are particularly difficult because plausible outputs can progressively erode accuracy and safety margins. Sensor diversity improves observability only when common dependencies are considered: multiple sensors may share clocks, calibration pipelines, processors, power supplies, environmental sensitivities, or training data. Sensor outputs should therefore expose health, freshness, provenance, and uncertainty, while disagreement should trigger risk-sensitive fusion or degraded operation rather than unconditional averaging~\citep{avizienis2004dependability,scholz2024sensor}.

Tactile and force sensing are particularly important in contact-rich tasks because vision alone cannot directly establish force, shear, slip, deformation, or jamming~\citep{tactile2025outlook,scholz2024sensor}. High-risk contact anomalies require a rapid local response, such as force limitation, controlled withdrawal, or stopping, without waiting for high-level reasoning. Local force and collision protection should remain independent of, and not directly overridable by, the task model.

\paragraph{Computation, communication, and energy.}
Embodied systems distribute computation across embedded controllers, real-time processors, accelerators, edge servers, and cloud services~\citep{macenski2022robot,casini2025survey}. Functions should be allocated according to latency, criticality, availability, and computational demand. High-level reasoning may tolerate variable latency, whereas balance, braking, force regulation, and collision response require predictable local execution.

Average latency is insufficient for safety-relevant functions. Analysis should cover the complete sensing-to-actuation path, including data freshness, tail latency, deadline misses, scheduling interference, network delay, and actuator response~\citep{sha1990priority,teper2024end,sobhani2023timing}. A numerically correct command may still be unsafe if executed too late.

Remote dependencies require explicit degraded behavior. Communication loss, stale authorization, or unavailable cloud services should revoke capabilities whose assumptions no longer hold, while local functions for balance, load support, braking, and minimum-risk transitions remain available during network degradation. Energy and thermal conditions likewise affect both nominal capability and fallback. Systems should retain sufficient power and thermal margin for critical sensing, control, communication, and braking, and task admission should account for the resources required for safe recovery rather than nominal completion alone.

\paragraph{Actuation and mechanical embodiment.}
Actuators convert commands into force and motion. Saturation, backlash, friction change, wear, overheating, brake failure, leakage, and encoder faults can invalidate planning and control assumptions. System-health estimates should therefore influence motion limits, task admission, and fallback selection.

Mechanical design can reduce failure consequences independently of perception, planning, and remote communication. Low inertia, compliant structures, rounded surfaces, protected pinch points, mechanical stops, clutches, series elasticity~\citep{pratt1995sea}, and passive safety mechanisms can limit force or preserve stable configurations. Active mechanisms, including impedance control~\citep{hogan1984impedance} and variable-stiffness robot skins~\citep{pang2020coboskin}, further support compliant physical interaction. Tools, payloads, and end effectors are part of the effective embodiment: changes to them alter inertia, workspace, sensing coverage, braking distance, contact pressure, and control authority. Such changes should update operating limits and trigger configuration-specific validation.

\subsection{Multi-Rate System Integration}
\label{subsec:multirate_integration}

Embodied systems operate across multiple timescales. Semantic reasoning and task planning typically run more slowly than policy generation, trajectory control, balance regulation, contact response, and hardware interlocks. A trustworthy architecture assigns each function an explicit update rate, authority boundary, input-freshness requirement, monitor, and fallback~\citep{casini2025survey}.

This separation prevents high-level models from acquiring unrestricted physical authority. A model may propose a grasp or trajectory, whereas local controllers enforce validated workspace, velocity, force, energy, and stability limits. They must also report withdrawal, slip, saturation, and protective stopping so that task-level reasoning does not continue from an invalid state.

Integration requires explicit cross-layer contracts in which time is a first-class property. Proposals should specify targets, reference frames, preconditions, uncertainty, active constraints, duration, completion criteria, and recovery options. Controllers should expose tracking error, actuator margin, contact state, stopping capability, and timing validity; sensors should expose health, calibration, freshness, and provenance. Because state estimates, trajectories, permissions, and safety margins are valid only for bounded intervals, maximum data age, command lifetime, monitor period, stopping time, and response latency should be explicit. Execution should be restricted when these bounds are violated rather than proceeding under stale assumptions~\citep{teper2024end,sobhani2023timing}.

\subsection{Fault Containment and Fallback}
\label{subsec:fault_containment_fallback}

Fault handling comprises detection, isolation, estimation, containment, degradation, and recovery~\citep{quamar2024faultdiagnosis,gao2015faultsurvey,sabry2024review,milecki2023review}. Model-based methods compare observed behavior with physical predictions; signal-based methods detect sensor and actuator anomalies; data-driven methods capture multivariate failure patterns; and hybrid methods combine physical structure with learned representations. All remain limited by fault coverage, detection delay, and diagnostic uncertainty.

Diagnosis is useful only when it triggers an appropriate physical response. Depending on the fault and remaining control authority, the system may compensate for a sensor, switch channels, reconfigure control, reduce speed, restrict workspace, disable a capability, unload an object, or transition to a minimum-risk condition~\citep{baioumy2021fault}. High-risk execution should not continue unchanged merely because fault classification remains uncertain.

Local protection provides the final containment barrier. Collision reflexes, force limits, safety filters, watchdogs, hardware interlocks, and fail-safe brakes should operate independently of general-purpose reasoning. Their end-to-end detection-and-response latency must remain below the interval over which a detected hazard becomes unrecoverable~\citep{ames2019cbf,haddadin2017collisions,cortez2021safe}.

Fallback and degradation are task- and embodiment-dependent. Immediate stopping may suit a stationary manipulator but be unsafe for a balancing robot, a platform carrying a person, or a robot holding a hot or fragile object. Depending on the remaining physical margin, a minimum-risk response may instead require controlled retreat, stable support, reduced-force execution, safe unloading, or short stabilization. When a reduced capability set remains useful and safe, the system may lower speed, restrict autonomy, disable sensor-dependent skills, or shrink its workspace. Recovery must restore valid task preconditions before nominal execution resumes; authorization to resume belongs to the deployment layer.

\subsection{Cyber-Physical and Distributed Execution}
\label{subsec:cyber_physical_execution}

Embodied systems often distribute perception, planning, control, and monitoring across robots, edge infrastructure, cloud services, and operator stations. Partial failure is therefore more common than total failure. Each unavailable or untrusted dependency should have a specified effect on executable capabilities, timing limits, and fallback behavior.

Cybersecurity violations can directly alter physical risk: sensor spoofing corrupts state estimation, timing attacks invalidate fusion, denial-of-service attacks delay protection, and compromised credentials grant access to physical capabilities~\citep{botta2023cyber}. Secure boot, authenticated communication, protected keys, integrity and anti-replay checks, network segmentation, and trustworthy logging consequently contribute to system assurance~\citep{rose2020zerotrust,mayoral2022sros2,deng2022security}.

Multi-robot systems introduce shared workspaces, resource conflicts, and common failure dependencies. Local collision avoidance and minimum-risk behavior should remain available under inconsistent communication or unavailable peers, and immediate physical protection should not depend on global consensus~\citep{lamport1982byzantine,deng2021investigation,ishii2022overview}.

\subsection{Hardware--Software Co-Design}
\label{subsec:hardware_software_codesign}

System assurance depends on coordinated physical design, computation, control, and model interfaces. Sensor placement should expose the state required for planning and diagnosis; compute allocation should preserve critical deadlines under peak load and degradation; mechanical design should limit the consequences of erroneous commands; and fault diagnosis should inform task admission and capability selection.

The central design principle is \emph{separation with coordination}. Task generation, physical execution, fast protection, and fault recovery should be sufficiently independent to constrain one another, while exchanging task intent, uncertainty, timing assumptions, system health, and safety constraints through explicit interfaces. No individual mechanism provides complete system assurance.

\subsection{Open Research Directions}
\label{subsec:system_open_directions}

\paragraph{End-to-end timing assurance.}
Developing worst-case response analysis that connects model inference, communication, control, and physical stopping margins across heterogeneous processors, networks, and dynamics~\citep{casini2025survey,teper2024end,sobhani2023timing}.

\paragraph{Fault coverage under open-world interaction.}
Extending diagnosis beyond predefined component faults to coupled faults, contact anomalies, configuration changes, cyber-physical attacks, and failures outside known diagnostic classes~\citep{sabry2024review,milecki2023review}.

\paragraph{Adaptive protection with bounded risk.}
Developing context-dependent physical envelopes that improve task performance while reverting to conservative fallback when the perception and health estimates on which they depend become unreliable~\citep{pang2020coboskin,cortez2021safe}.

\paragraph{Assurance-preserving reconfiguration.}
Determining which assumptions and supporting evidence are invalidated by tool changes, hardware replacement, controller updates, or capability transfer, and regenerating the affected evidence efficiently.
\section{Evidence-Level Trustworthiness}
\label{sec:evidence_layer}

The evidence layer determines which claims about an embodied system are justified. Its purpose is not to rank systems or report benchmark scores, but to establish whether the evaluated system supports sufficient task capability and acceptable residual risk within a specified deployment boundary; a credible claim must identify the system configuration, task distribution, embodiment, environment, authority structure, evaluation conditions, and assumptions under which it holds. Evaluation mainly establishes evidence before deployment, after which operational observations may strengthen, restrict, or invalidate the claim while the deployment layer governs data collection, intervention, and revalidation.

\subsection{Evaluation Objects and Claims}
\label{subsec:evaluation_objects}

Evaluation begins by defining the object of the claim, which may be a perception model, planner, policy, safety monitor, controller, hardware component, integrated robot, multi-robot system, or complete human-robot application. Evidence does not transfer automatically between objects: a calibrated perception model does not establish timely intervention, a verified controller does not establish the validity of task intent, and reliable components do not imply reliable integration under shared dependencies.

Evaluation strength should match the intended claim. Component tests isolate specific mechanisms, whereas integrated-system tests capture interactions among perception, planning, timing, control, hardware, and human behavior~\citep{ashmore2019assuring,seshia2022toward}; deployment-oriented claims require evidence from the configured system, since changes in sensors, tools, payloads, controllers, networks, or authority alter both capability and risk~\citep{hawkins2021guidance,burton2022safety}.

\subsection{Joint Evaluation of Capability and Safety}
\label{subsec:joint_evaluation}

Capability and safety must be evaluated jointly. Capability metrics include task completion, quality, efficiency, latency, throughput, robustness, and generalization; safety metrics include collisions, near misses, constraint violations, force or pressure exceedance, unsafe duration, minimum separation, hazardous object states, and exposure to people or assets; supporting metrics include reliability, degradation, recovery time, intervention delay, unnecessary rejection, and availability~\citep{ray2019benchmarking,ji2023safety}.

The primary joint outcome is the \emph{safe-success rate}:
\begin{equation}
    \mathrm{SSR}
    =
    \frac{
        N\!\left(
        \text{task completed}
        \land
        \text{no unacceptable violation}
        \right)
    }{
        N(\text{evaluated trials})
    }.
\end{equation}

This unweighted trial fraction should be interpreted together with its component outcomes:
\begin{itemize}[leftmargin=*]
    \item \textbf{Safe success:} the task is completed without an unacceptable violation;
    \item \textbf{Safe failure:} the task is not completed, but safety and recoverability are preserved;
    \item \textbf{Unsafe success:} the task is completed through behavior that violates a safety, procedural, semantic, or authority constraint;
    \item \textbf{Unsafe failure:} the task fails and also produces unacceptable risk or harm.
\end{itemize}

These outcomes separate limited capability from unsafe behavior and excessive conservatism; safe failure should be further split into justified intervention and unnecessary interruption, since frequent refusal preserves safety yet can make the system operationally ineffective. Because averages cannot characterize tail risk or severity, evaluation should report scenario-conditioned results, exposure, confidence intervals where appropriate, violation severity, worst observed outcomes, and intervention burden, so that a high-energy contact and a minor safety-filter activation are not weighted equally.

\subsection{Complementary Forms of Evidence}
\label{subsec:evidence_types}

Evidence differs in rigor, realism, and validity boundary (Table~\ref{tab:evidence_types}); four complementary forms are particularly relevant.

\paragraph{Formal evidence.}
Temporal specifications, reachability bounds, invariant sets, and barrier certificates establish properties under explicit models and assumptions~\citep{ames2019cbf,bansal2017reachability}.

\paragraph{Statistical evidence.}
Confidence intervals, calibrated uncertainty~\citep{guo2017calibration,lakshminarayanan2017ensembles,snoek2019can}, conformal risk bounds~\citep{angelopoulos2021conformal}, and reliability estimates support probabilistic claims under a specified sampling process~\citep{koh2021wilds}.

\paragraph{Empirical evidence.}
Datasets, benchmarks, simulation, fault-injection campaigns, and physical experiments record integrated behavior that is difficult to formalize.

\paragraph{Operational evidence.}
Deployment records of task outcomes, interventions, failures, near misses, recovery, and environmental change offer high ecological validity but remain observational, and an absence of recorded harm may reflect limited exposure, conservative use, or chance rather than acceptable risk~\citep{kalra2016driving}.

\begin{table*}[th]
\centering
\small
\setlength{\tabcolsep}{8pt}
\renewcommand{\arraystretch}{1.25}

\caption{Complementary evidence types and their validity boundaries.}
\label{tab:evidence_types}

\begin{tabularx}{\textwidth}{
    p{0.14\textwidth}
    p{0.4\textwidth}
    X
}
\toprule
\textbf{Evidence}
&
\textbf{What it supports}
&
\textbf{Main boundary}
\\
\midrule

Formal
&
Verified properties under explicit models
&
Model and assumption fidelity
\\

Statistical
&
Probabilistic risk and reliability claims
&
Sampling and distribution validity
\\

Empirical
&
Observed behavior under controlled tests
&
Test and scenario coverage
\\

Operational
&
Behavior under real deployment exposure
&
Observed exposure and reporting coverage
\\

\bottomrule
\end{tabularx}
\end{table*}

No evidence type is sufficient in isolation; assurance should combine sources through explicit claims and assumptions rather than aggregate heterogeneous results into a single score.

\subsection{Staged Evaluation}
\label{subsec:staged_evaluation}

Evaluation should raise realism, exposure, and physical consequence gradually,
from scalable repeatable exploration to embodiment, contact, timing, human
interaction, hardware degradation, and organizational conditions
(Table~\ref{tab:evaluation_stages}).

\begin{table*}[t]
\centering
\small
\setlength{\tabcolsep}{5pt}
\renewcommand{\arraystretch}{1.3}

\caption{Evaluation coverage across stages of trustworthy embodied-system
assessment. A checkmark denotes primary coverage, a triangle denotes
supporting or partial coverage, and a dash denotes no direct coverage.}
\label{tab:evaluation_stages}

\begin{tabularx}{\textwidth}{
    p{0.25\textwidth}
    >{\centering\arraybackslash}X
    >{\centering\arraybackslash}X
    >{\centering\arraybackslash}X
    >{\centering\arraybackslash}X
    >{\centering\arraybackslash}X
    >{\centering\arraybackslash}X
}
\toprule
\textbf{Evaluation stage}
&
\textbf{Requirements}
&
\textbf{Model behavior}
&
\textbf{Closed loop}
&
\textbf{Timing \& integration}
&
\textbf{Physical interaction}
&
\textbf{Field exposure}
\\
\midrule

Specification \& formal analysis
&
$\checkmark$
&
$\triangle$
&
$\triangle$
&
$\triangle$
&
--
&
--
\\

Offline model evaluation
&
$\triangle$
&
$\checkmark$
&
--
&
$\triangle$
&
--
&
--
\\

Simulation \& digital twins
&
$\triangle$
&
$\checkmark$
&
$\checkmark$
&
$\triangle$
&
$\triangle$
&
--
\\

Software / hardware in the loop
&
$\triangle$
&
$\triangle$
&
$\checkmark$
&
$\checkmark$
&
$\triangle$
&
--
\\

Controlled physical evaluation
&
$\triangle$
&
$\triangle$
&
$\checkmark$
&
$\checkmark$
&
$\checkmark$
&
--
\\

Bounded pilot deployment
&
$\triangle$
&
$\triangle$
&
$\checkmark$
&
$\checkmark$
&
$\checkmark$
&
$\checkmark$
\\

\bottomrule
\end{tabularx}
\end{table*}

The process is iterative rather than sequential: physical failures should
update simulation scenarios, virtual testing should prioritize physical
experiments, and deployment observations should trigger regression testing,
with advancement between stages requiring evidence that the claim survives the
increased realism and exposure. Virtual proving grounds efficiently generate
combinations of objects, people, lighting, contact properties, communication
delays, hardware degradation, and malicious inputs that are costly or unsafe
to reproduce physically~\citep{muratore2022robot,fremont2020formal}. Scalable randomized
simulation~\citep{chen2025robotwin,Mu_2025_CVPR}, deformable- and garment-oriented
environments~\citep{wang2025dexgarmentlab}, and long-horizon procedural
benchmarks such as laboratory automation~\citep{lan2025autobio} broaden
scenario coverage, while standardized simulation-to-reality infrastructure
connects these results to physical execution~\citep{chen2026robodojo}.
Coverage should be defined over capability- and risk-relevant dimensions
rather than episode count.

Learned world models extend virtual proving grounds to action-conditioned
prediction and can support prospective capability and risk evaluation, but
their task-progress and hazard predictions require calibration, uncertainty
bounds, and validation against physical
trajectories~\citep{tang2026safedojo,chen2026robodojo}. Simulation is itself
part of the evidence chain: its physics, sensing, contact, latency, human
behavior, and failure models must be validated against physical outcomes
~\citep{deitke2020robothor,li2024evaluating},
since visual similarity alone does not establish that task success, contact
behavior, or risk estimates transfer.

Physical evaluation remains indispensable for embodiment-, contact-, and
timing-dependent behavior~\citep{heo2025furniturebench}. Distributed and standardized real-robot evaluation
platforms~\citep{atreya2025roboarena,yakefu2025robochallenge} improve
reproducibility and exposure across sites and embodiments, although physical
coverage and duration stay limited relative to simulation.

\subsection{Stress, Failure, and Long-Horizon Evaluation}
\label{subsec:stress_failure_evaluation}

Nominal testing is insufficient, since trustworthiness also depends on behavior under uncertainty, faults, and adversarial conditions. Fault injection can introduce sensor bias, timestamp error, processing delay, actuator degradation, network loss, low power, thermal throttling, and software failure~\citep{hsueh1997faultinjection}, and evaluation should measure not only whether failure occurs but whether it is detected, contained, and followed by appropriate degradation or fallback.

Adversarial evaluation should examine prompt injection, conflicting instructions, visual patches, deceptive content, sensor spoofing, replayed commands, compromised dependencies, and attempts to bypass physical safeguards~\citep{li2026safetyembodied,ma2026breaks}, using bounded environments and protective procedures so that physical red teaming does not itself create unacceptable risk. Automated scenario generation can search for small safety margins, sensor disagreement, excessive uncertainty, timing violations, poor recoverability, and monitor failures more efficiently than random sampling~\citep{riedmaier2020scenario,dreossi2019verifai,fremont2019scenic,koren2018adaptive,wang2021advsim,tian2018deeptest}, and generated cases should be grouped by causal mechanism to avoid mistaking repeated variants for broad coverage.

Long-horizon evaluation must examine complete execution traces rather than final states alone~\citep{kim2026longhorizon}, through precondition satisfaction, subtask ordering, accumulated risk, warning lead time, intervention timing, recovery attempts, irreversible state changes, and repeated retries; sustained runs are also needed to expose sensor drift, mechanical wear, thermal effects, memory accumulation, and changing human behavior. Safeguards must be part of the evaluated object, assessed by detection coverage, false intervention, response latency, physical effectiveness, recovery outcome, computational overhead, and effect on useful task capability.

\subsection{Benchmarks, Reproducibility, and Transfer}
\label{subsec:benchmarks_transfer}

Existing benchmarks offer standardized tasks but often emphasize binary completion or aggregate reward~\citep{james2020rlbench,yu2020metaworld,zhu2020robosuite,mees2022calvin,liu2023libero,gu2023maniskill2,li2024behavior}. 
Across both embodied tasks and related perception domains, standardized benchmarks increasingly support unified evaluation and reproducible comparison~\citep{ma2024imdl}; recent manipulation benchmarks further broaden task, scene, and embodiment coverage~\citep{atreya2025roboarena,chen2026rmbench,yang2026eventvla,chen2026manipulationnetinfrastructurebenchmarkingrealworld}, while visuo-tactile benchmarks extend evaluation to contact-rich perception that vision alone cannot assess~\citep{chen2026univtac}. 
Trustworthiness-oriented benchmarks should additionally specify the embodiment, sensor and action interfaces, hazard taxonomy, safety constraints, scenario distribution, evaluation budget, exposure, and statistical procedures, and should report safe success, unsafe success, violation severity, intervention burden, efficiency, and availability.

Reproducibility requires versioned configurations and execution traces, since firmware, control gains, model checkpoints, camera placement, tools, payloads, network conditions, random seeds, and human procedures can materially change results; independent evaluation matters most when developers select scenarios or interpret ambiguous success and violation criteria.

Cross-embodiment transfer requires caution: semantic knowledge may transfer across robots, while geometry, action scaling, payload, compliance, sensing, braking, contact dynamics, and safety thresholds remain embodiment-dependent, so task transfer must be distinguished from evidence that safety and recovery behavior remain valid on a new platform.

\subsection{Assurance Cases}
\label{subsec:assurance_cases}

An assurance case connects a bounded deployment claim to supporting arguments, assumptions, and evidence~\citep{kelly2004gsn}. Its top-level claim should identify the system and model versions, task distribution, embodiment and tools, operating domain, authority structure, capability requirements, and acceptable-risk criteria, while subclaims address model behavior, physical limits, fault containment, cybersecurity, recovery, human interaction, and deployment monitoring. The argument should show how formal, statistical, empirical, and operational evidence jointly supports the claim, which assumptions are shared, and where uncertainty remains; evidence should not be pooled without considering dependence, common-cause failure, and mismatched validity boundaries.

The output of the evidence layer should state:
\begin{itemize}[leftmargin=*]
    \item what the configured system is supported to do;
    \item under which tasks and operating conditions the claim applies;
    \item which capability and safety criteria have been established;
    \item which evidence and assumptions support the claim;
    \item which changes or observations require restriction or revalidation.
\end{itemize}

This output grounds deployment decisions: the deployment layer then monitors whether the system remains within the supported boundary and triggers intervention or revalidation when the claim is no longer justified.

\subsection{Open Research Directions}
\label{subsec:evidence_open_directions}

\paragraph{Risk-oriented scenario coverage.}
Coverage measures that capture causal diversity, exposure, consequence severity, and uncertainty rather than trial count, and that establish whether tested scenarios span the dominant sources of physical and procedural risk.

\paragraph{Compositional evidence.}
Determining when component-level guarantees support integrated-system claims and when shared assumptions or interface failures invalidate composition.

\paragraph{Simulation-to-reality validity.}
Quantifying which task and safety properties transfer from simulation, world models, and digital twins to physical systems, rather than treating realism as a single visual measure.

\paragraph{Long-term and cross-embodiment evaluation.}
Infrastructure for sustained operation, diverse embodiments, and contact-rich tasks with reproducible failure analysis and configuration-specific validity boundaries.
\section{Deployment-Level Trustworthiness}
\label{sec:deployment_layer}

\begin{figure*}[t]
    \centering
    \includegraphics[width=0.9\textwidth]{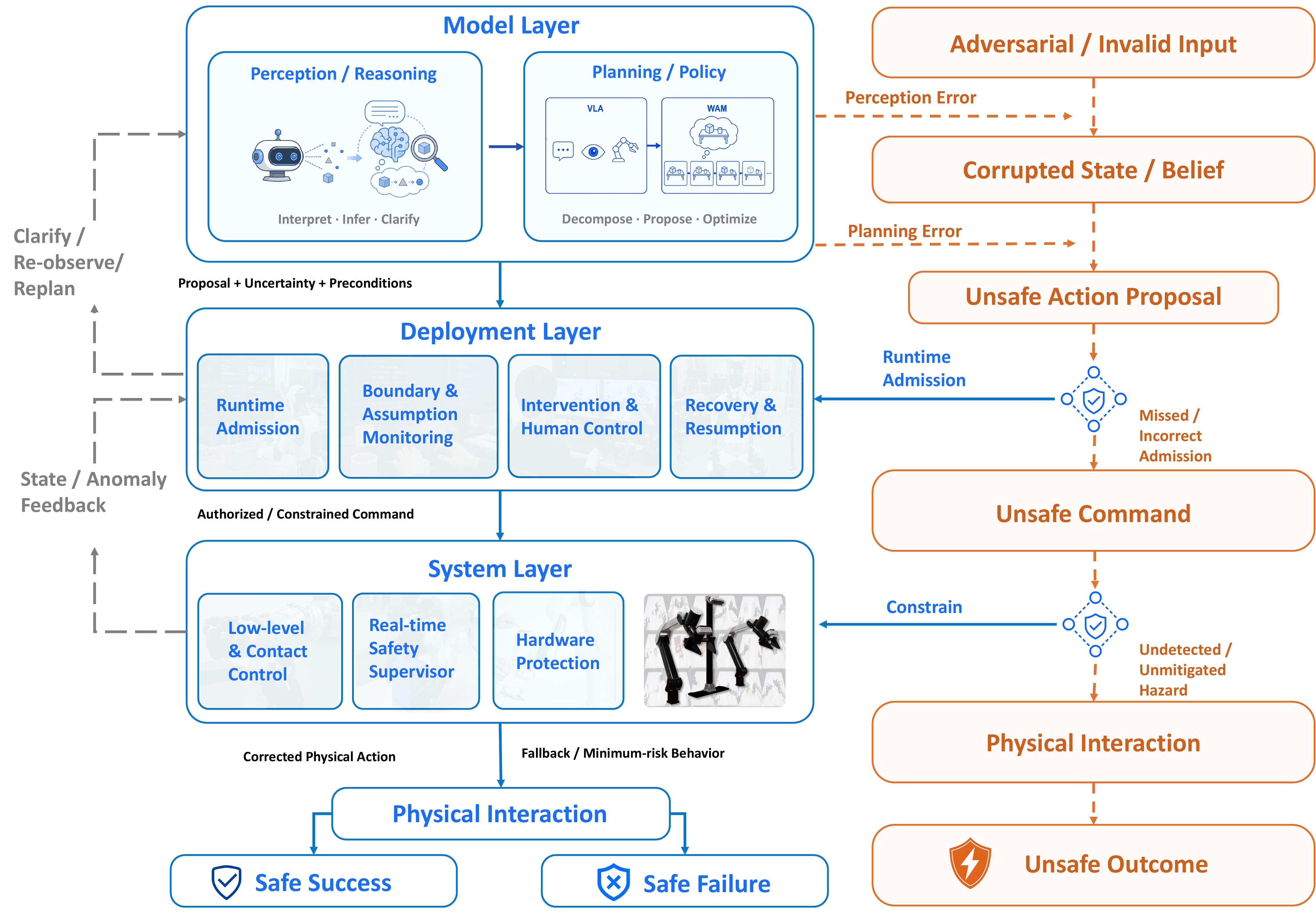}
    \caption{Runtime governance between model-level action proposals and physical execution. The deployment layer performs runtime admission, monitors boundaries and assumptions, coordinates intervention and human control, and governs recovery and resumption, while system-level safeguards constrain residual hazards during physical interaction.}
    \label{fig:runtime_governance}
\end{figure*}

The deployment layer maintains the validity of a trustworthiness claim during use and separates model-level action proposal from execution authority. A system may satisfy its predeployment evaluation criteria yet become unsupported when tasks, environments, hardware, software, users, or procedures change. This layer therefore determines whether execution remains within its supported boundary and whether deviations trigger timely restriction, intervention, recovery, or revalidation. Whereas the system layer provides physical protection, fault containment, and fallback, the deployment layer governs their operational use through authority definition, assumption monitoring, intervention coordination, incident management, and safety-relevant change control. Figure~\ref{fig:runtime_governance} presents the resulting runtime-governance process.

As illustrated in Figure~\ref{fig:runtime_governance}, state-estimation errors may lead to unsafe action proposals. Runtime admission can prevent proposals that violate known constraints from becoming executable commands~\citep{alshiekh2018shielding}, while system-level safeguards provide a further barrier before physical interaction. If these barriers fail, are unavailable, or act too late, the resulting hazard may propagate to an unsafe outcome. Trustworthiness therefore depends on containing errors across multiple layers rather than assuming correct model behavior alone.

\subsection{Operational Boundaries and Runtime Admission}
\label{subsec:runtime_admission}

A deployment claim applies to a bounded combination of system version, task, embodiment, environment, authority, and evidence. Its boundary should specify the permitted robot configuration, tools, payloads, object classes, workspace, contact modes, human proximity, communication conditions, task types, and supervision. Broad labels such as ``indoor operation'' or ``service robotics'' are insufficient because they omit variables that directly affect capability and risk.

The operational design domain from autonomous driving provides a useful precedent~\citep{iso21448,sae2021j3016,koopman2019odd}. An embodied-system boundary must additionally represent task semantics, physical interaction, tool use, and embodiment-specific dynamics: validation on one gripper, payload, controller, or workspace does not establish support for deployment after these conditions change.

Runtime admission determines whether a requested capability may execute under current conditions, verifying:
\begin{itemize}[leftmargin=*]
    \item task and capability compatibility;
    \item system identity, configuration, and health;
    \item environmental and interaction conditions;
    \item required permissions and human supervision;
    \item availability of monitoring, protection, and fallback;
    \item validity of the evidence supporting the requested execution.
\end{itemize}

Decisions may allow, restrict, deny, or escalate execution. A capability validated in an empty workspace may require reduced speed near a person, additional approval for a hazardous tool, or denial after a safety-relevant sensor failure. Admission is therefore not a one-time startup check: its supporting assumptions must remain monitored throughout execution.

\subsection{Runtime Monitoring}
\label{subsec:deployment_monitoring}

Runtime monitoring assesses whether the assumptions supporting admission and deployment remain valid. Relevant signals include task progress, model uncertainty, distribution shift, system health, timing, active constraints, human proximity, communication status, resource availability, and environmental change~\citep{qin2026runtimegovernance,hendrycks2017ood,rahman2021monitoring,sinha2022ood}.

Monitoring must connect to actionable criteria, since anomaly detection is ineffective if the response occurs after safe intervention becomes infeasible~\citep{leucker2009runtime,luckcuck2019formal}. Each monitored condition should specify:
\begin{itemize}[leftmargin=*]
    \item the assumption or constraint being monitored;
    \item the signal and update rate used to assess it;
    \item the threshold or decision rule indicating loss of validity;
    \item the required response and its deadline;
    \item the authority responsible for confirming or resolving the event.
\end{itemize}

Monitoring should address both sudden violations and gradual degradation. Repeated replanning, rising uncertainty, prolonged task duration, frequent safety-filter activation, declining grasp quality, or increasing intervention frequency can indicate shrinking safety margins even when no individual event constitutes a failure~\citep{chandola2009anomaly}. Monitors can themselves fail through incomplete coverage, stale information, incorrect thresholds, service interruption, or alert overload. Their failure behavior must therefore be specified, and safety-critical local protection must remain available when centralized monitoring or remote services do not.

\subsection{Authority, Intervention, and Human Control}
\label{subsec:authority_intervention}

Deployment governance separates task generation from execution authority: models may propose behavior, but the deployed system determines which actions are permitted under the current task, state, and policy~\citep{sheridan1992telerobotics,parasuraman2000model}. Least privilege restricts each model, user, service, and robot to the capabilities required by its role~\citep{saltzer1975protection}.

Intervention may modify, pause, stop, replan, restrict, or transfer execution. The response should be scaled to the remaining physical margin and consequence severity: minor uncertainty may warrant re-observation or reduced speed, whereas loss of trusted state, control authority, or safety-critical sensing may require immediate fallback. Rapid physical hazards remain the responsibility of automatic local protection because human response is too slow for many contact, balance, and collision events~\citep{haddadin2017collisions}. Humans are better suited to ambiguity, authorization, ethical or organizational judgment, recovery selection, and high-consequence decisions that permit sufficient response time.

Human control must be effective rather than nominal: an emergency button or takeover interface provides little protection when the operator lacks information, training, physical access, or time~\citep{santonidesio2018mhc,bainbridge1983ironies}. A handover should communicate the current state, detected hazard, active task, system limitations, available actions, and remaining response time~\citep{endsley1995situation}. Takeover latency and outcome quality should be evaluated under realistic workload and communication conditions~\citep{eriksson2017takeover}. Authority transfer must be explicit, recording who or what controlled execution, when authority changed, what information supported the transfer, and whether the recipient assumed control. Otherwise, ambiguous shared control may leave each party expecting the other to act.

\subsection{Degradation, Recovery, and Resumption}
\label{subsec:deployment_recovery}

When nominal execution is no longer justified, the deployment layer selects among restriction, controlled degradation, fallback, recovery, and termination; the system layer realizes these decisions through local control and hardware. A trustworthy outcome need not entail task completion: when safe completion is no longer possible, controlled failure is preferable to an unsafe outcome.

Graceful degradation preserves a reduced capability set, for example by lowering speed, restricting workspace, disabling a sensor-dependent skill, requiring continuous supervision, or limiting execution to low-risk tasks. The degraded mode must have an explicit capability boundary rather than be treated as nominal operation with reduced performance. Because immediate stopping can itself be unsafe for a robot that is balancing, supporting a person, or holding a hazardous object, the minimum-risk condition is task- and embodiment-dependent, ranging from stopping, stable support, controlled retreat, safe unloading, and reduced-force behavior to transfer to a human~\citep{zanella2026phri,isots15066}.

Recovery should restore valid task and system preconditions before execution resumes, potentially through re-observation, replanning, regrasping, recalibration, component replacement, or human inspection. Resumption should occur only after the condition is resolved and authority is renewed. Repeated automated recovery attempts should be bounded to avoid accumulated contact, wear, delay, or human burden.

\subsection{Incident Response and Operational Evidence}
\label{subsec:incident_response}

Deployment monitoring should record more than accidents. Near misses, unsafe successes, unexpected stops, repeated retries, abnormal contact, constraint activations, human corrections, degraded modes, and failed recovery can reveal weaknesses before serious harm occurs~\citep{favaro2017examining}. Logs should identify system and model versions, hardware configuration, active task, relevant observations, proposed and executed actions, authorization decisions, interventions, timing, and outcome. Synchronized timestamps and configuration provenance support reconstruction of cross-layer failure propagation, subject to proportionate access, retention, and privacy controls.

Incident response should first stabilize the situation and protect people and assets, then preserve evidence, identify technical and organizational causes, correct the system or procedure, and validate the correction before return to service~\citep{leveson2011safer}. Responsibility should not be assigned solely to the component producing the final visible error, since model assumptions, hardware conditions, integration, supervision, and organizational decisions may all contribute.

Operational observations provide evidence under actual exposure but do not independently establish safety: an absence of recorded harm may reflect limited use, conservative operators, incomplete reporting, or chance~\citep{kalra2016driving}. Field evidence should therefore be interpreted together with exposure, predeployment evaluation, and known monitoring coverage.

\subsection{Change Control and Revalidation}
\label{subsec:change_control}

A deployment claim can lapse without an obvious failure. Relevant changes include:
\begin{itemize}[leftmargin=*]
    \item model fine-tuning, replacement, or online adaptation;
    \item controller, middleware, firmware, or dependency updates;
    \item sensor replacement or recalibration;
    \item changes in tools, payloads, mechanical configuration, or workspace;
    \item network, cloud-service, or cybersecurity changes;
    \item new tasks, users, interaction modes, or authority policies;
    \item incidents, newly discovered hazards, or expanded operating domains.
\end{itemize}

Each safety-relevant change should undergo impact analysis. Minor changes may require targeted regression testing, whereas changes to embodiment, physical authority, hazard exposure, or core behavior may require renewed system evaluation. Updates should not enter deployment merely because they improve average benchmark performance. Online adaptation requires particular care because field data may improve performance while introducing behavioral drift or weakening validated constraints~\citep{sculley2015hidden,lesort2020continual}. Updated models should remain versioned and traceable, with evidence that both capability and residual risk remain acceptable.

The resulting assurance process forms a closed loop~\citep{kephart2003autonomic}:
\begin{equation}
\begin{aligned}
    \text{validate}
    &\rightarrow \text{deploy}
    \rightarrow \text{monitor}
    \rightarrow \text{diagnose} \\
    &\rightarrow \text{restrict or update}
    \rightarrow \text{revalidate}.
\end{aligned}
\end{equation}

The loop does not require complete recertification after every observation. Rather, it connects changes and incidents to the specific claims, assumptions, and tests they affect.

\subsection{Open Research Directions}
\label{subsec:deployment_open_directions}

\paragraph{Runtime boundary recognition.}
Combining semantic, physical, system-health, and authority information to detect, with sufficient lead time, when open-world conditions leave the validated domain.

\paragraph{Proportionate intervention.}
Developing responses between continued autonomy and full shutdown that minimize risk while preserving useful task capability and recoverability.

\paragraph{Effective human authority.}
Establishing measurable conditions for observability, comprehension, reaction time, workload, and successful transfer, particularly for long-horizon and contact-rich tasks.

\paragraph{Cross-organizational responsibility.}
Developing technical interfaces and accountability mechanisms that coordinate incident response and change control across model providers, manufacturers, integrators, deployers, maintainers, operators, and users.
\section{Levels of Trustworthy Embodied Intelligence}
\label{sec:levels_deployment}

Binary labels such as trustworthy or untrustworthy provide limited guidance for embodied systems. The strength of a trustworthiness claim depends on the scope of the claimed capability, the severity of potential consequences, the system safeguards in place, the supporting evidence, and the governance maintained during use~\citep{brundage2020toward}. We therefore introduce six levels of Trustworthy Embodied Intelligence (TEI), from T0 to T5, representing progressively stronger and more sustained forms of justified trustworthiness for embodied-system deployment.

The proposed levels are:
\begin{itemize}[leftmargin=*]
    \item \textbf{T0: No Trustworthy Embodied Intelligence;}
    \item \textbf{T1: Constrained Trustworthy Embodied Intelligence;}
    \item \textbf{T2: Partially Trustworthy Embodied Intelligence;}
    \item \textbf{T3: Conditionally Trustworthy Embodied Intelligence;}
    \item \textbf{T4: Highly Trustworthy Embodied Intelligence;}
    \item \textbf{T5: Sustained Trustworthy Embodied Intelligence.}
\end{itemize}

These labels characterize the strength and scope of a substantiated trustworthiness claim, rather than an intrinsic or universal property of a robot~\citep{brundage2020toward}. A level applies to a bounded combination of:
\begin{equation}
    \left\langle
    \text{system version},
    \text{task},
    \text{embodiment},
    \text{operating domain},
    \text{authority},
    \text{evidence}
    \right\rangle,
\end{equation}
not to a model family, platform, or product in isolation. The same robot may therefore occupy different levels across tasks, tools, environments, configurations, and supervision conditions.

The hierarchy is analytical rather than a certification scheme, universal risk scale, or autonomy taxonomy. Domain-specific standards must define the quantitative thresholds, acceptable residual risk, required evidence, and independent assessment needed to substantiate each level.

\subsection{Assessment Dimensions}
\label{subsec:trustworthiness_dimensions}

A TEI level is assessed across five jointly necessary dimensions, extending multi-dimensional accounts of trustworthy AI~\citep{li2023trustworthy} to embodied deployment:
\begin{enumerate}[leftmargin=*]
    \item \textbf{Capability adequacy:} required task success, quality, efficiency, robustness, and generalization;
    \item \textbf{Safety and residual risk:} unsafe success, violation severity, exposure, and foreseeable consequences within acceptable bounds;
    \item \textbf{System assurance:} sensing, computation, control, hardware protection, fault containment, degradation, and fallback for dependable physical execution;
    \item \textbf{Evidence sufficiency:} formal, statistical, empirical, and operational evidence for the claimed system, task, embodiment, and operating domain;
    \item \textbf{Deployment governance:} authority, monitoring, intervention, incident response, change control, and revalidation that maintain the claim during use.
\end{enumerate}

High capability cannot compensate for uncontrolled physical risk, nor can restrictive safeguards compensate for failure to provide the intended service. A TEI level therefore cannot exceed the least adequately supported safety-critical dimension.

\begin{figure*}[t]
    \centering
    \includegraphics[width=0.8\textwidth]{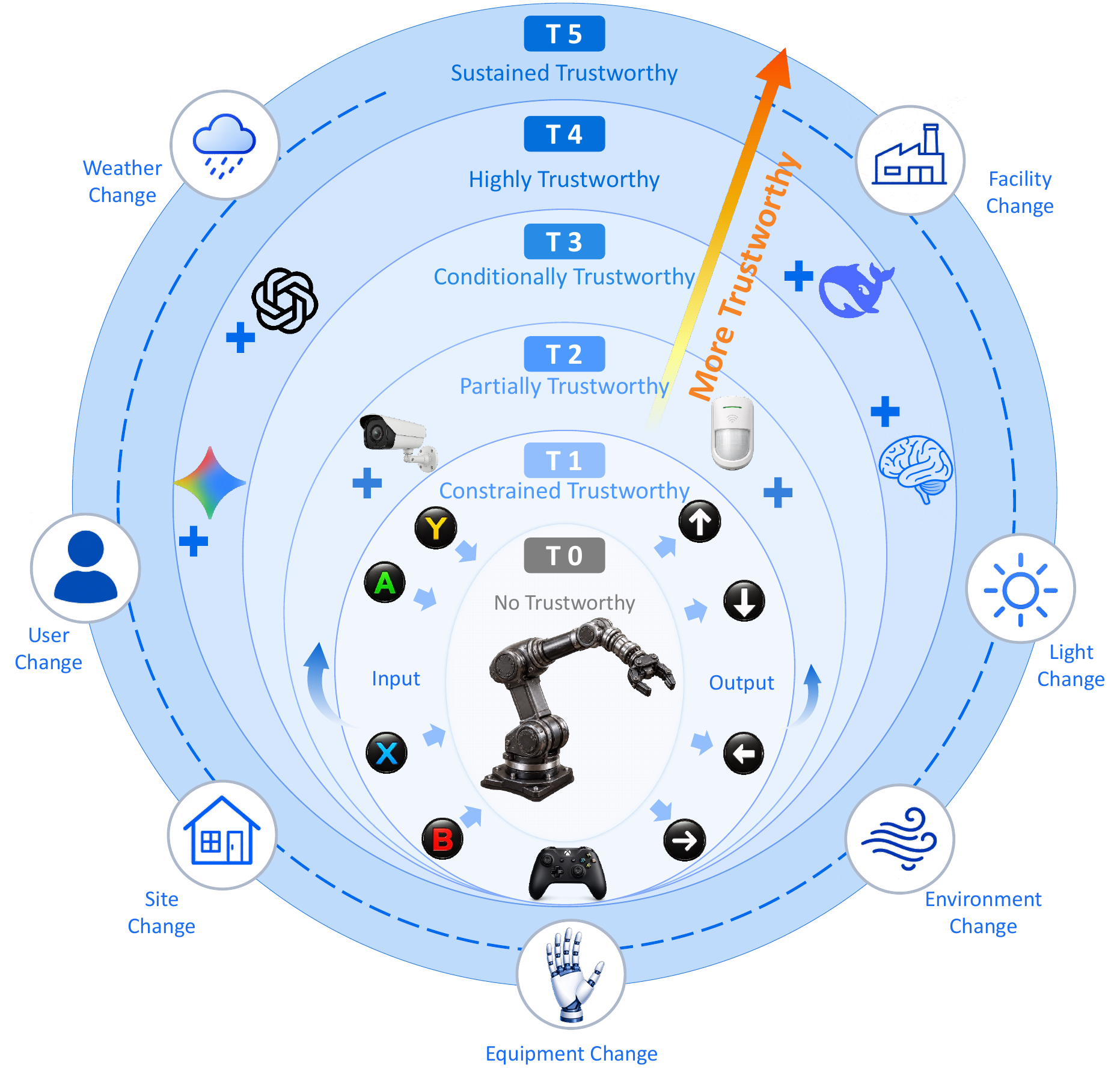}
    \caption{Proposed T0--T5 levels of Trustworthy Embodied Intelligence. Higher levels represent progressively stronger and more sustained trustworthiness claims, with T5 emphasizing maintenance of a claim under governed changes in users, sites, equipment, environments, and operating conditions.}
    \label{fig:trustworthiness_levels}
\end{figure*}

Figure~\ref{fig:trustworthiness_levels} summarizes the cumulative progression from T0 to T5.

The concentric structure represents increasing strength and scope of the substantiated trustworthiness claim, rather than increasing model capability alone. In particular, the outer progression emphasizes that T5 concerns maintaining a justified boundary as deployment conditions evolve; it does not assume that trustworthiness automatically generalizes to every new condition.

\subsection{Level Definitions}
\label{subsec:level_definitions}

The six levels characterize progressively stronger forms of justified trustworthiness. Requirements are cumulative, and each level remains bounded to the system, task, embodiment, operating domain, authority structure, and evidence for which its claim has been established.

\paragraph{T0: No Trustworthy Embodied Intelligence.}
At T0, no substantiated trustworthiness claim has been established for deployment. Concepts, prototypes, and selective demonstrations may exhibit promising capability, but task selection, resets, teleoperation, human assistance, failure frequency, or unsafe intermediate behavior may be unreported or insufficiently evaluated. Reports should disclose the tested configuration, assistance, and known failures. The absence of visible harm should not be interpreted as evidence of safety.

\paragraph{T1: Constrained Trustworthy Embodied Intelligence.}
At T1, trustworthiness is substantiated only for narrow and largely predefined behavior under explicit physical constraints. Independent limits on position, speed, force, energy, workspace, and emergency stopping prevent software from acquiring unrestricted physical authority. Humans supervise continuously and remain responsible for abnormal conditions. A guarded industrial sequence in a restricted workspace is illustrative. T1 supports trustworthy execution only within explicitly constrained behavior; it does not imply broad adaptation or autonomous recovery.

\paragraph{T2: Partially Trustworthy Embodied Intelligence.}
At T2, trustworthiness extends beyond predefined behavior to selected or composed bounded skills, but only part of the system's autonomous capability is sufficiently supported for trustworthy execution. These skills remain observable and governable through explicit authority boundaries and runtime monitoring of task state, system health, timing, and known faults. The system can stop or enter a controlled condition after a recognized deviation, but timely human fallback remains part of the assurance strategy. T2 therefore supports partial trustworthy autonomy rather than responsibility for a complete task loop.

\paragraph{T3: Conditionally Trustworthy Embodied Intelligence.}
At T3, trustworthiness extends from individual skills to responsibility for a complete task loop under explicitly specified and validated conditions. The system may perform perception, interpretation, planning, execution, verification, and recovery from specified deviations within a defined operational boundary.

The system must detect when its state, uncertainty, or assumptions no longer justify continued execution and must degrade, recover, or request assistance before safe continuation becomes infeasible. A warehouse robot performing specified pick-and-place tasks within validated workspaces, payloads, and object classes is illustrative. The resulting claim does not automatically transfer to new tools, contact modes, public environments, or untested object classes.

\paragraph{T4: Highly Trustworthy Embodied Intelligence.}
At T4, a high degree of trustworthiness is substantiated across multiple tasks or skill combinations within a validated operating domain. The system can autonomously handle a defined set of foreseeable failures without depending on immediate human response.

Loss of a planner, learned model, communication service, or noncritical sensor must not disable local protection. The system should select among replanning, degraded execution, safe unloading, controlled retreat, or termination according to the remaining physical margin. T4 may support extended operation within a carefully specified domain when the supporting evidence and accepted residual risk are adequate; it implies neither competence nor safety outside that domain.

\paragraph{T5: Sustained Trustworthy Embodied Intelligence.}
At T5, justified trustworthiness can be sustained as tasks, configurations, models, and operating conditions evolve. The defining property is not universal trustworthiness, but a governed process for determining whether existing evidence continues to justify the system's current authority.

The deployment process monitors the validity of relevant assumptions and retains, restricts, or revokes capabilities accordingly. Adaptation and system evolution are governed through traceable updates, behavioral regression analysis, operational evidence, incident feedback, and repeated assurance review~\citep{batarseh2021survey}.

T5 therefore denotes sustained maintenance of a bounded trustworthiness claim under controlled evolution. It does not imply universal safety, unrestricted autonomy, or competence for arbitrary tasks, embodiments, environments, or conditions.

\subsection{Transition Logic}
\label{subsec:level_transition}

Because requirements are cumulative, moving to a higher TEI level requires evidence that additional capability or operational responsibility is matched by stronger system assurance, supporting evidence, and deployment governance. Improvements in model architecture, benchmark performance, task generality, or autonomy alone do not justify progression.

The principal transitions are:
\begin{itemize}[leftmargin=*]
    \item \textbf{T0 to T1:} from demonstrated capability without a substantiated trustworthiness claim to reproducible trustworthy behavior under explicit physical constraints;

    \item \textbf{T1 to T2:} from narrowly constrained predefined behavior to partially trustworthy autonomous skill execution supported by runtime monitoring, explicit authority boundaries, and defined responses to known failures;

    \item \textbf{T2 to T3:} from trustworthiness of individual skills or portions of autonomous capability to conditional trustworthiness of a complete task loop within an explicitly validated boundary;

    \item \textbf{T3 to T4:} from conditional trustworthiness that may rely on timely human fallback to autonomous handling of defined foreseeable failures within the validated domain;

    \item \textbf{T4 to T5:} from high trustworthiness for a relatively stable validated configuration to sustained trustworthiness under governed changes in tasks, models, configurations, and operating conditions.
\end{itemize}

A system may also be downgraded when evidence expires, monitoring becomes unavailable, unresolved incidents occur, or its configuration or operating domain changes. TEI levels therefore characterize the currently justified strength of a bounded trustworthiness claim rather than permanent system maturity.

\subsection{Relation to Existing Taxonomies}
\label{subsec:relation_existing_taxonomies}

\begin{table*}[t]
\centering
\caption{Illustrative comparison of AgiBot G1--G5, SAE L0--L5, and the proposed T0--T5 Trustworthy Embodied Intelligence levels. Rows show their respective progressions only; the levels are not directly equivalent.}
\label{tab:level_comparison}
\renewcommand{\arraystretch}{1.35} 
\resizebox{\textwidth}{!}{%
\begin{tabular}{lll}
\toprule
\textbf{AgiBot G1--G5}
&
\textbf{SAE L0--L5}
&
\textbf{Proposed TEI T0--T5}
\\
\midrule

\textemdash
&
\textbf{L0} No driving automation
&
\textbf{T0} No Trustworthy Embodied Intelligence
\\

\textbf{G1} Basic automation
&
\textbf{L1} Driver assistance
&
\textbf{T1} Constrained Trustworthy Embodied Intelligence
\\

\textbf{G2} Reusable atomic skills
&
\textbf{L2} Partial driving automation
&
\textbf{T2} Partially Trustworthy Embodied Intelligence
\\

\textbf{G3} Data-driven skill learning
&
\textbf{L3} Conditional driving automation
&
\textbf{T3} Conditionally Trustworthy Embodied Intelligence
\\

\textbf{G4} General operation model
&
\textbf{L4} High driving automation
&
\textbf{T4} Highly Trustworthy Embodied Intelligence
\\

\textbf{G5} End-to-end embodied model
&
\textbf{L5} Full driving automation
&
\textbf{T5} Sustained Trustworthy Embodied Intelligence
\\

\bottomrule
\end{tabular}%
}
\end{table*}

Although the three hierarchies use similar numerical progressions, they classify fundamentally different properties, and the row alignment in Table~\ref{tab:level_comparison} is illustrative only. AgiBot G1--G5 describes the evolution of embodied capability from basic automation toward increasingly general embodied models, and related roadmaps likewise track skill integration, generality, and task capability~\citep{limou2026embodiedaction,xspark2026waic}. 
SAE L0--L5 classifies driving automation according to the allocation of the dynamic driving task and fallback responsibility~\citep{sae2021j3016}. The proposed T0--T5 hierarchy instead asks how strongly, and over what scope, the trustworthiness of an embodied capability can be justified and maintained; a TEI level is determined jointly by the five dimensions of Section~\ref{subsec:trustworthiness_dimensions} rather than by intelligence or autonomy alone.

Consequently, a system may use a highly general model yet remain at T1 or T2 if monitoring, fallback, evidence, or change control are insufficient. Conversely, a narrowly capable industrial system may support a T4-level trustworthiness claim within a tightly constrained and extensively validated domain.

\subsection{Implications for Standardization}
\label{subsec:level_standardization}

The T0--T5 TEI hierarchy provides a common structure for comparing bounded trustworthiness claims, but practical adoption requires domain-specific profiles. Each profile should define:
\begin{itemize}[leftmargin=*]
    \item the task and operating-domain boundary;
    \item quantitative capability and safety thresholds;
    \item mandatory system safeguards and fallback functions;
    \item required evaluation scenarios and evidence strength;
    \item monitoring, intervention, and incident-reporting obligations;
    \item conditions for reassessment, downgrade, suspension, or renewal.
\end{itemize}

Such profiles allow the hierarchy to support benchmarking, procurement, staged deployment, and future standards without assuming that a single universal threshold is appropriate for all embodied systems.
\section{Synthesis of Prior Work}
\label{sec:prior_work_summary}

The preceding sections reviewed mechanisms within the model, system, evidence, and deployment layers. This section synthesizes their historical development, relative maturity, and cross-layer distribution. Its purpose is not to rank methods, but to identify where reusable foundations have been established, where recent research activity concentrates, and where end-to-end assurance remains incomplete.

\subsection{Evolution of the Research Landscape}
\label{subsec:research_evolution}

Research relevant to trustworthy embodied intelligence has developed through several overlapping streams.

\paragraph{Foundations in physical and system assurance.}
Early work established principles that remain central to trustworthy physical execution: impedance control, compliant actuation, collision analysis, dependable computing, fault diagnosis, and real-time scheduling~\citep{avizienis2004dependability,haddadin2017collisions,sha1990priority,pratt1995sea,hogan1984impedance}. These foundations support the system layer by controlling physical interaction, containing component faults, and preserving timing properties. They remain necessary when task decisions are generated by learned models.

\paragraph{Safe learning and runtime constraint enforcement.}
From approximately 2015, research examined how learned policies could optimize task performance subject to safety constraints through safe reinforcement learning, reachability analysis, barrier functions, and predictive safety filters~\citep{ames2019cbf,garcia2015saferl,wabersich2021psf,bansal2017reachability}. In parallel, autonomous driving advanced scenario-based evaluation, safety of the intended functionality, operational design domains, and fallback responsibility~\citep{riedmaier2020scenario,iso21448,sae2021j3016}. These developments linked model-, system-, and evidence-level mechanisms, although most guarantees remained bounded by specific dynamics, hazards, and operating conditions.

\paragraph{General-purpose embodied models.}
Since 2022, language-grounded planning, vision--language--action models, world-model-based methods, and cross-embodiment datasets have expanded the range of tasks that robots can interpret, plan, and execute~\citep{brohan2023rt2,kim2024openvla,black2024pi0,hafner2023dreamerv3,ahn2022saycan,o2024open,firoozi2025foundation}. This progress has shifted emphasis toward generality, semantic reasoning, and transfer, while exposing new trustworthiness challenges: hallucinated affordances, incomplete physical grounding, opaque action interfaces, long-horizon error propagation, and uncertain transfer of safety evidence across embodiments~\citep{li2024embodied}.

\paragraph{Lifecycle-oriented trustworthiness.}
Recent work increasingly treats trustworthiness as a system and lifecycle problem, addressing physical risk, long-horizon safety, runtime governance, fault-tolerant robotics, tactile interaction, and integrated simulation and physical evaluation~\citep{tactile2025outlook,kojima2025physicalrisk,li2026safetyembodied,ma2026breaks,kim2026longhorizon,qin2026runtimegovernance,quamar2024faultdiagnosis,paterson2025safety}. However, this cross-layer perspective remains less consolidated than research on model capability and local physical safety.

\subsection{Future Research Roadmap and Standardization Alignment}
\label{subsec:future_roadmap}

The remaining challenges span multiple technical and organizational scales. To relate the open problems identified in Sections~\ref{sec:model_layer}--\ref{sec:deployment_layer}, we organize future research across five levels: infrastructure and data, foundation models, trustworthy learning, evaluation and verification, and physical-intelligence applications. This research-oriented stack complements, rather than replaces, the four-layer system framework.

\begin{figure*}[t]
    \centering
    \includegraphics[width=\textwidth]{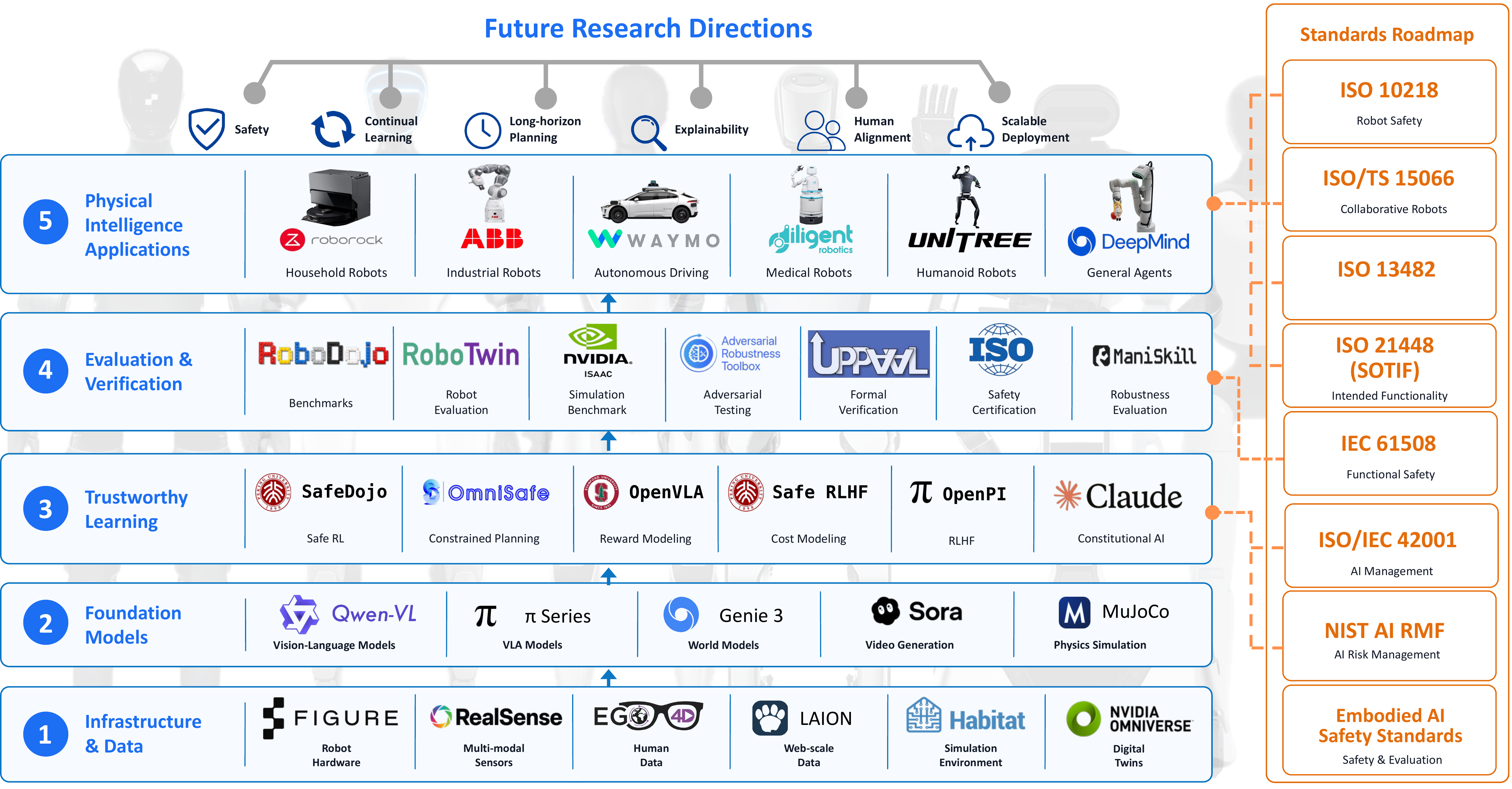}
    \caption{Future research roadmap for trustworthy embodied intelligence, spanning infrastructure and data, foundation models, trustworthy learning, evaluation and verification, and physical-intelligence applications, with cross-cutting research directions and alignment with existing safety and AI-governance standards.}
    \label{fig:future_research_roadmap}
\end{figure*}

As illustrated in Figure~\ref{fig:future_research_roadmap}, progress at higher levels depends on foundations established below. Trustworthy applications require not only capable foundation models and learning mechanisms, but also reliable infrastructure, representative data, and systematic evaluation and verification. Safety, continual learning, long-horizon planning, explainability, human alignment, and scalable deployment cut across these levels rather than belonging to any single component.

The standards roadmap in Figure~\ref{fig:future_research_roadmap} indicates that existing robot-safety, functional-safety, intended-functionality, and AI-governance standards already address important parts of this stack~\citep{iso21448,iso10218,iec61508,nist2023airmf}. However, no single framework spans the full combination of learned intelligence, physical interaction, evaluation, and lifecycle governance. Future standardization should therefore connect these foundations while developing assurance requirements specific to embodied AI.
\section{Conclusion}
\label{sec:conclusion}

Embodied intelligence translates computational decisions into physical motion, contact, and environmental change; task capability alone therefore cannot justify deployment. We define trustworthy embodied intelligence as \emph{sustained safe success}: reliable completion of intended tasks while residual risk remains within acceptable bounds. Any resulting trustworthiness claim is necessarily bounded to a particular system configuration, task, embodiment, operating domain, authority structure, and body of evidence. Achieving such a claim requires four interdependent layers---model, system, evidence, and deployment---none of which is sufficient in isolation because assumptions, uncertainty, and failures propagate across their interfaces.

The literature reveals an uneven landscape. Model capability has advanced rapidly through foundation models, world models, and generalist policies, while system assurance draws on mature foundations in control, mechanics, real-time systems, and dependable computing. However, the composition of these foundations with learned components remains comparatively underexplored, as do claim-oriented evidence, runtime governance, effective human intervention, incident learning, and assurance under change. To structure comparison, we introduce a T0--T5 hierarchy of Trustworthy Embodied Intelligence. The hierarchy characterizes progressively stronger and more sustained forms of bounded trustworthiness, assessed jointly by capability, safety, system assurance, evidence, and governance rather than by intelligence or autonomy alone. It is not a certification scheme and requires domain-specific thresholds, procedures, and acceptable-risk criteria.

Progress will require stronger cross-layer contracts, physically grounded uncertainty, recovery-centered execution, risk-oriented evaluation, configuration-aware revalidation, and measurable human and organizational assurance. Autonomous-driving practice offers useful precedents for bounded operating domains, scenario-based validation, fallback, and staged deployment, but these principles must be adapted to diverse embodiments, intentional contact, open-ended instructions, and task-specific recovery. Trustworthiness implies neither universal competence nor the absence of failure. Rather, it requires the sustained alignment of capability, physical authority, evidence, and operational responsibility throughout the lifecycle: a system whose useful capabilities are bounded, supported, monitored, and revised as its configuration and environment change.

\section*{\XRGradientText{Contributions}}

\colorlet{xrLeft}{AbsBlue}
\colorlet{xrMid}{AbsBlue!50!AbsCyan}
\colorlet{xrRight}{AbsCyan}
\phantomsection
\label{contributions}
\begingroup

\makeatletter
\newcommand\blfootnote[1]{%
  \begingroup
    \renewcommand\thefootnote{}%
    \renewcommand\@makefntext[1]{\noindent##1}%
    \footnote{#1}%
    \addtocounter{footnote}{-1}%
  \endgroup
}
\makeatother
\newcommand{\aff}[1]{\textsuperscript{#1}}
\newcommand{\xrrole}[1]{{\large\bfseries #1\par}\vspace{0.3em}}

\setlength{\columnsep}{0.04\textwidth}


\noindent
\begin{minipage}[t]{0.47\textwidth}
  \vspace{0pt}
  \raggedright\nohyphens

  \xrrole{Project Leaders}
  Xinyu Yang\aff{1,2}, Tianxing Chen\aff{1,3}

  \vspace{1em}

  \xrrole{Corresponding Authors}
  Wenbo Ding\aff{1,2}, Ping Luo\aff{3}, Qi Xiong\aff{1,2}

  \vspace{1em}

  \xrrole{Core Contributors}
  Honghao Su\aff{1}, Chenze Yu\aff{1}, Zhangzheng Tu\aff{5},
  Minxuan Wang\aff{1}, Yue Chen\aff{1,4}, Yuxiao Huo\aff{1,2}
\end{minipage}
\hfill
\begin{minipage}[t]{0.47\textwidth}
  \vspace{0pt}
  \raggedright\nohyphens

  \xrrole{Contributors}
  Lingfeng Zhang\aff{2}, Yan Huang\aff{2}, Yan Qin\aff{1,6},
  Shaolong Zhu\aff{1}, Qiwei Liang\aff{1,6}, Hekun Tian\aff{1,2},
  Shujia Liu\aff{1,2}, Guangyu Chen\aff{1,2}, Junhao Gong\aff{1},
  Zixuan Li\aff{1}, Wenwei Lin\aff{1}, Zijian Lin\aff{1},
  Wenxuan Zhu\aff{1,2}, Eric J Chen\aff{1,2}, Yue Yuan\aff{1},
  Qize Yu\aff{4}, Jiaqi Liang\aff{4}, Haowen Yan\aff{1},
  Hengfei Zhao\aff{1,2}, Weijie Wan\aff{1}, Zikun Xiao\aff{1},
  Junyuan Tang\aff{1}, Baijun Chen\aff{1}, Kai-Chong Lei\aff{1,2},
  Kaixuan Wang\aff{3}, Kailun Su\aff{1}, Zanxin Chen\aff{5}, Yao Mu\aff{5},
  Renjing Xu\aff{6}, Chuqiao Lyu\aff{1,2}
\end{minipage}

\blfootnote{%
 Author affiliations:
  \textsuperscript{1}Xspark AI\quad
  \textsuperscript{2}THU\quad
  \textsuperscript{3}MMLab@HKU\quad
  \textsuperscript{4}PKU\quad
  \textsuperscript{5}SJTU\quad
  \textsuperscript{6}HKUST (GZ)%
}

\endgroup

\clearpage
\bibliographystyle{unsrtnat}
\bibliography{main}

\clearpage
\beginappendix

\section{Suggested Literature Search Protocol}
\label{app:search_protocol}

Future updates of this review may use a structured search across IEEE Xplore, ACM Digital Library, Web of Science, Scopus, Google Scholar, and arXiv. Search strings should combine terms related to trustworthy embodied AI, physical-intelligence safety, vision--language--action safety, runtime assurance, fault-tolerant robot control, physical human--robot interaction, world-model-based safe reinforcement learning, virtual proving grounds, robot-safety benchmarks, and continuous assurance. Backward and forward reference tracing should complement database search for foundational work in control, dependability, networking, and standards.

For reproducibility, each search should record the database, query string, search date, date range, document type, screening decision, and reason for exclusion. Results should be deduplicated before title--abstract screening and full-text assessment.

Inclusion criteria should identify whether a work provides a direct trustworthiness mechanism, an evaluation or benchmark, a foundational method with clear embodied relevance, a standard or assurance process, or a documented deployment or incident case. Capability-oriented papers should be included only when they establish a system interface, expose a risk pressure, or define an evidence boundary relevant to the analysis. Company materials may illustrate architectures or deployment claims, but should be labeled as first-party evidence unless supported by technical publications or independent evaluation.

\section{Literature Extraction Template}
\label{app:extraction_template}

For each paper, standard, benchmark, or deployment case, extract:
\begin{itemize}[leftmargin=*]
    \item bibliographic metadata and publication status;
    \item embodiment, task, tool, environment, and human role;
    \item trustworthiness objective and explicit claim;
    \item initiating risk or failure mode;
    \item originating layer and cross-layer propagation path;
    \item planning-, policy-, execution-, or contact-time intervention;
    \item lifecycle phase: predeployment, runtime, or post-incident;
    \item mechanism, required models, and assumptions;
    \item evidence type, metrics, scenarios, duration, and baselines;
    \item supported deployment claim and validity boundary;
    \item known limitations, residual risk, and revalidation triggers.
\end{itemize}

\section{Compact Terminology}
\label{app:terminology}

\begin{table}[h]
\centering
\small
\caption{Working terminology used in this review.}
\label{tab:terminology}
\begin{tabular}{p{0.19\linewidth}p{0.79\linewidth}}
\toprule
\textbf{Term} & \textbf{Working meaning} \\
\midrule
Physical intelligence & Intelligence whose decisions are grounded in and realized through a physical system \\
Embodied intelligence & An agent whose perception, cognition, learning, and action are shaped by its body and environment \\
Trustworthy embodied intelligence & Sustained safe success within a bounded claim, supported by capability, safety, reliability, controllability, governability, evaluability, and deployability \\
Governability & Ability to authorize, constrain, observe, interrupt, recover, audit, and transfer execution \\
ODD & Specified operating conditions within which a function or system is intended to operate \\
Safe success & Task completion without an unacceptable safety, procedural, semantic, or authority violation \\
Minimum-risk condition & Context-dependent condition that limits further harm when nominal operation is no longer justified \\
Assurance case & Structured claims, assumptions, arguments, and evidence supporting a bounded deployment claim \\
\bottomrule
\end{tabular}
\end{table}

\section{Review Maintenance and Evidence Updates}
\label{app:planned_completion}

The current review is a structured synthesis of the supplied seed literature and project materials rather than a complete systematic review. Future updates should broaden coverage of dependable computing, operating systems, networking, control, autonomous-driving assurance, documented deployment experience, and primary standards.

Internal and industry materials should remain explicitly identified as first-party sources and should not be treated as independent validation. As new literature, evaluations, or operational evidence emerge, the review should update its cross-layer analysis, validity boundaries, evidence tables, and figures while preserving traceability between each revised claim and its supporting sources.

\end{document}